\documentclass[final,5p,times,twocolumn]{elsarticle}

\usepackage{amssymb}
\usepackage{amsmath}
\usepackage{caption}
\usepackage{booktabs}
\usepackage{multirow}
\usepackage{pifont}
\usepackage{enumitem}
\usepackage{xcolor}
\usepackage{url}
\newcommand{\modified}[1]{\textcolor{black}{#1}}

\begin{document}

\begin{frontmatter}



\title{A Synergistic CNN-Transformer Network with Pooling Attention Fusion for Hyperspectral Image Classification}


\author[label1]{Peng Chen} %% Author name
\author[label2,label1]{Wenxuan He}
\author[label3]{Feng Qian}
\author[label4]{Guangyao Shi}
\author[label5,label1]{Jingwen Yan\corref{cor1}}
\affiliation[label1]{organization={College of Engineering},%Department and Organization
            addressline={Shantou University}, 
            city={Shantou},
            postcode={515063}, 
            state={Guangdong},
            country={China}}

\affiliation[label2]{organization={School of Electronic Information and Communications},
            addressline={Huazhong University of Science and Technology}, 
            city={Wuhan},
            postcode={430074}, 
            state={Hubei},
            country={China}}
\affiliation[label3]{organization={Changchun Institute of Optics, Fine Mechanics and Physics},
            addressline={Chinese Academy of Sciences}, 
            city={Changchun},
            postcode={130033}, 
            state={Jilin},
            country={China}}
\affiliation[label4]{organization={School of Computer Science and Technology},
            addressline={Chongqing University of Posts and Telecommunications}, 
            city={Chongqing},
            postcode={400065}, 
            country={China}}
\affiliation[label5]{organization={School of Intelligent Manufacturing and Electrical Engineering},
            addressline={Guangzhou Institute of Science and Technology}, 
            city={Guangzhou},
            postcode={510000}, 
            state={Guangdong},
            country={China}}
\cortext[cor1]{Corresponding author. \\E-mail address: jwyan@stu.edu.cn (J. Yan).}
%% Abstract
\begin{abstract}
In the hyperspectral image (HSI) classification task, each pixel is categorized into a specific land-cover category or material. Convolutional neural networks (CNNs) and transformers have been widely used to extract local and non-local features in HSI classification. Recent works have utilized a multi-scale vision transformer (ViT) to enhance spectral feature capture and yield promising results. However, most existing methods still face challenges in the effective joint use of spatial-spectral information and in preserving information across layers during the propagation process. To address these issues, we propose a synergistic CNN-Transformer network with pooling attention fusion for HSI classification, which collaboratively utilizes CNNs and ViT to process spatial and spectral features separately. Specifically, we propose a Twin-Branch Feature Extraction (TBFE) module, which employs 3D and 2D convolution in parallel to comprehensively extract spectral and spatial features from HSI. A hybrid pooling attention (HPA) module is designed to aggregate spatial attention. Moreover, a cascade transformer encoder is employed for global spectral feature extraction, and a simple yet efficient cross-layer feature fusion (CFF) module is designed to reduce the loss of crucial information in the previous network layers. Extensive experiments are conducted on several representative datasets to demonstrate the superior performance of our proposed method compared to the state-of-the-art works. Code is available at \url{https://github.com/chenpeng052/SCT-Net.git}
\end{abstract}

%% Keywords

\end{frontmatter}



\section{Introduction}

Hyperspectral images (HSI) are composed of multiple contiguous and narrow spectral bands, each of which serves to identify different materials or land-cover categories \cite{9143471,10231043}. In contrast to RGB and multispectral images, HSI offers more detailed spectral information \cite{10360846}. Consequently, the HSI data finds wide applications in urban environmental planning, geological hazard monitoring, and precision agriculture \cite{10172252}. HSI classification is a crucial task in the field of remote sensing \cite{9328201}, involving the accurate classification and identification of ground objects by assigning specific category labels for each pixel \cite{chen2025synergistic}. This process enhances surface feature recognition accuracy and supports informed decision-making and management, thereby improving resource utilization efficiency and promoting environmental protection.

In the early phases of HSI classification research, most approaches primarily focused on the spectral features of the images \cite{8697135}. \modified{Classical machine learning techniques, such as support vector machine (SVM) \cite{1323134}, k-nearest neighbors (KNN) \cite{5555996}, and random forest (RF) \cite{1396322}, were widely adopted due to their effectiveness in handling high-dimensional data. However, these methods frequently lacked comprehensive spectral property analysis of HSI and underutilized feature correlations among pixels in the spatial domain. } Some approaches have introduced diverse mathematical morphology operators to extract spatial features from HSI effectively. The morphological profile (MP) \cite{4686022} enhances local details and texture information in images, while the extended morphological profile (EMP) \cite{1396321} incorporates supplementary structural elements and operations to enhance local feature descriptions. The extended multiattribute profile (EMAP) \cite{5664759} integrates information from multiple attributes for a more comprehensive image feature depiction, thereby boosting accuracy in image analysis and classification. Nonetheless, these approaches frequently encounter challenges in capturing intricate data patterns and interactions.

\modified{To mitigate redundancy in high spectral resolution data, prevalent strategies have employed statistical techniques to extract critical band information from HSI and discard less significant bands\cite{10713459}. Independent Component Analysis (ICA) \cite{CAO2003321} separates spectral sources by representing input data as mutually independent components, allowing it to preserve more physical information. However, ICA suffers from high computational complexity and is sensitive to noise. Linear Discriminant Analysis (LDA) \cite{6617430} selects a feature subspace that effectively distinguishes between different classes by calculating class scatter matrices, thereby enhancing the data's suitability for classification tasks. However, its performance is constrained by dimensionality reduction, and it assumes that the data follows a normal distribution, which is often difficult to satisfy in practical applications. Autoencoders \cite{10371361}, as a nonlinear dimensionality reduction technique, automatically learn low-dimensional representations of the data and can capture complex nonlinear relationships~\cite{chen2026wmoe}. However, autoencoders require a large number of parameters for training and are sensitive to the data distribution. In contrast, Principal Component Analysis (PCA) \cite{1420303}, a widely used linear dimensionality reduction technique, extracts the most important information from the data by maximizing variance, effectively reducing the data’s dimensionality while being computationally efficient and independent of the data distribution. Although the aforementioned dimensionality reduction methods inevitably cause some loss of spectral information in HSI data, PCA, compared to other techniques, not only effectively removes redundancy in HSI but also preserves the spatial information of the spectrum without degradation, making it more suitable for HSI dimensionality reduction tasks.}

The rapid advancement of deep learning technologies has significantly driven innovation in HSI processing techniques \cite{10574887}, leading to the development of numerous classification methods \modified{aimed at extracting} deeper features\cite{chen2026dyc}. Chen et al. \cite{7018910} introduced an innovative image classification framework based on a deep belief network (DBN) that integrates spectral-spatial feature extraction and classification. However, this framework requires converting the input training image blocks into 1D feature vectors, altering the spatial information of the original images. Zhao et al. \cite{7450160} combined 2D convolution and max pooling to propose SSFC, a classification framework where convolutional neural networks (CNNs) automatically identify high-level spatial features. These features are then fused by combining spectral and spatial information. With the widespread adoption of CNNs, researchers have investigated various CNN-based architectures for HSI classification. Yue et al. \cite{yue2015spectral} were among the first to introduce a hybrid approach that combines PCA with deep convolutional neural networks (DCNNs), presenting a feature map generation algorithm to produce spectral and spatial feature maps for hierarchical deep feature extraction. Chakraborty et al. \cite{chakraborty2021spectralnet} integrated wavelet transform with CNNs to extract enhanced spectral features, forming a spatial-spectral feature vector for classification. Shi et al. \cite{10384612} proposed using a large-sized spectral convolution kernel for the spectral dimension of the hyperspectral cube, which facilitates downsampling and feature extraction while addressing the issue of missing local information in large-scale tokens.

The introduction of 3D convolution has significantly advanced the utilization of spectral-spatial information in HSI classification. Chen et al. \cite{chen2016deep} integrated regularization techniques into a 3D CNN-based feature extraction model, \modified{effectively addressing the challenges of imbalanced high dimensionality and limited training samples in HSI classification.} Roy et al. \cite{8736016} combined features from both 3D and 2D CNNs to propose a hybrid spectral CNN, achieving promising performance with fewer parameters. Additionally, the high dimensionality of HSI data presents a challenge in preserving multi-scale spectral features. Zhou et al. \cite{10210615} innovatively integrated 3D CNN with a feature pyramid structure, employing multi-scale convolutional extraction to capture spectral-spatial features. However, gradient vanishing and explosion tend to occur with increased 3D convolutional layers. To address these challenges, Yang et al. \cite{9641863} proposed a cross-spectral attention component that utilizes \modified{both} local and global spectral information to generate band weights, effectively suppressing redundant spectral bands. Wang et al. \cite{10319785} developed a large kernel sparse ConvNet (LSCNet) weighted by multi-frequency attention, employing two parallel rectangular convolutional kernels to approximate a large kernel, \modified{expanding} the receptive field and \modified{enhancing} the model's ability to learn and capture features.

\modified{Substantial research has been conducted into other deep learning backbone architectures aimed at enhancing the extraction of spatial-spectral features from hyperspectral data \cite{10398446}.} Recurrent neural networks (RNNs) \cite{7914752}, a form of feedforward neural network, have been explored for HSI classification tasks. RNNs effectively process hyperspectral pixels as sequential data, facilitating the classification of information categories through network inference. \modified{Generative adversarial networks (GANs), which typically comprise a generative network and a discriminative network in competition, have proven their versatility for HSI classification tasks.} Zhu et al. \cite{8307247} proposed a robust 3D-GAN as a spectral-spatial classifier to enhance \modified{model generalization} and mitigate the sparse nature of training samples in HSI data. Moreover, graph convolutional networks (GCNs) have demonstrated promising results in HSI classification tasks. Ding et al. \cite{ding2022multi} introduced a novel multi-feature fusion network capable of extracting multi-scale pixel-wise local features for HSI classification. \modified{To reduce model complexity, Mamba \cite{gu2023mamba}, a novel structured state-space model, has shown great potential in training and inference. Huang et al. \cite{rs16132449} proposed the Spectral-Spatial Mamba model, which employs a dual-branch Mamba structure to separately extract spectral and spatial features, significantly enhancing the representation capability for HSI.} These backbone networks provide diverse ideas and insights for taking full advantage of HSI data characteristics.

In recent years, the transformer \cite{vaswani2017attention} has garnered significant research interest in natural language processing (NLP), characterized by its self-attention mechanism, which captures long-term dependencies by processing inputs into a series of tokens. Building on this success, the vision transformer (ViT) \cite{dosovitskiy2020image} represents a major advancement in visual tasks, including remote sensing. Given the extensive spectral bands in HSI, which contain highly correlated and granular information, it is essential to extract global dependencies to learn more refined features. Consequently, leveraging the ViT to capture spatial-spectral features in the HSI classification task can lead to superior performance. Hong et al. \cite{9627165} proposed a pure transformer-based network to aggregate spectral sequence information by embedding groups of neighboring bands. Despite its strong performance, this model ignored the spatial information in HSI. Sun et al. \cite{9684381} proposed to capture high-level spectral and spatial features through tokenization, addressing the limitations of CNN-based methods in extracting deep semantic features. Mei et al. \cite{9895238} introduced a grouped pixel embedding module to confine multi-head self-attention within the local spatial-spectral context. \modified{Shu et al. \cite{SHU2024107351} proposed a dual-attention transformer that embeds local spatial information into global spectral features, thus enhancing joint classification performance. Sun et al. \cite{10506482} proposed incorporating memory tokens into the multi-head attention mechanism to reduce information loss during the propagation process.}

Although existing methods have demonstrated remarkable effectiveness, the intrinsic high-dimensional characteristics and the challenge of effectively integrating multi-dimensional features pose significant obstacles to the HSI classification task. CNN-based methods are highly effective at analyzing spatial features but struggle to adequately leverage the sequential spectral information intrinsic to HSI. \modified{Conversely, transformer-based methods excel at identifying long-range dependencies between spatial and spectral data, but they encounter difficulties in extracting multi-dimensional features from the shallow features obtained through convolution, potentially leading to inadequate feature extraction. Additionally, while Mamba-based methods effectively reduce computational complexity, the interaction between different encoding layers is crucial. Relying solely on simple concatenation may weaken the connectivity between these layers.} To address these challenges, this article proposes a synergistic CNN-Transformer network with pooling attention fusion for HSI classification. The proposed method combines the local feature extraction capabilities of CNNs with the global modeling strengths of the transformer, effectively capturing and integrating discriminative features across both spatial and spectral dimensions. Specifically, our method employs two consecutive twin-branch convolution modules to extract spectral-spatial features from HSI, followed by a hybrid pooling attention module that reweights global information and further aggregates features from parallel branches through cross-dimensional interaction. To efficiently utilize features from various layers, a cross-layer feature fusion module is introduced to combine features from previous layers before feeding them into the subsequent encoder. Finally, a fully connected layer is used to recognize the extracted features and assign a land cover category to each pixel in the HSI. The main contributions of this article are summarized as follows:
\begin{itemize}
    \item A twin-branch feature extraction (TBFE) module is proposed, which can adjust channels and simultaneously leverage the advantages of both 2D and 3D convolution to effectively extract spatial and spectral features from HSI.  
    \item To establish local cross-channel interactions within each parallel branch, a hybrid pooling attention (HPA) module is introduced to integrate the output feature maps from the two parallel networks through a cross-spatial learning approach. 
    \item A simple yet efficient feature fusion cross-layer feature fusion (CFF) is proposed, which enhances information interaction between layers and reduces information loss by concatenating the features from different encoding layers.
    \item Extensive experiments on \modified{five} representative datasets verify that our method achieves satisfactory results compared with the state-of-the-art methods. 
\end{itemize}

\section{Materials and methods}
\label{sec2}
\begin{figure*}
    \centering
    \includegraphics[width=0.9\linewidth]{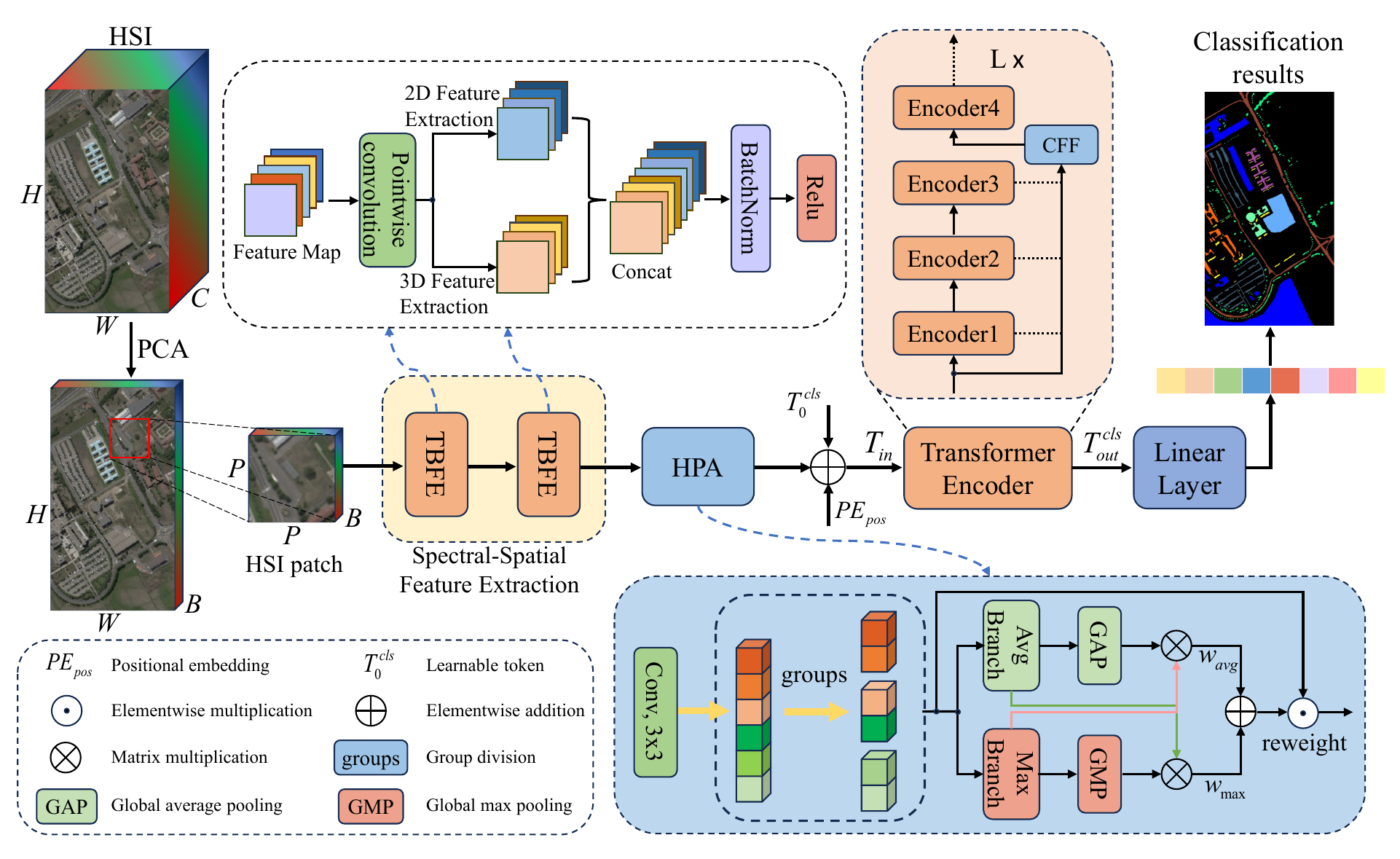}
    \caption{An overview illustration of the proposed synergistic CNN-Transformer network for the HSI classification task. The proposed network consists of several well-designed modules, i.e., TBFE, HPA, and CFF, making it better applicable for synergistic CNN-Transformer architecture.}
    \label{fig1}
\end{figure*}

In this section, we describe the overall structure of the HSI classification process using the proposed model, as illustrated in Fig.~\ref{fig1}. Initially, the HSI data preprocessing based on PCA is employed to reduce the dimensionality of the original HSI data. Subsequently, the reduced-dimensional data is processed through a spectral-spatial feature extraction module to extract and represent HSI features effectively. Then, a hybrid pooling attention module is proposed to obtain spectral attention weights. Finally, the weighted tokens are classified using the transformer encoder with cross-layer feature fusion. The classification results provide the predicted labels for the central pixels of the original patches.

\subsection{HSI Data Preprocessing}

Given the initial HSI data, denoted as $\mathit{I}\in\mathbb{R}^{H\times W\times C}$, where $H\times W$ represents the spatial dimensions and $C$ indicates the number of spectral channels. HSI data typically contains hundreds of spectral channels, with each pixel in $\mathit{I}$ associated with a specific material. \modified{However,} due to the limited availability of training samples, directly training a deep learning model increases the parameter count and significantly heightens the risk of overfitting. To mitigate this issue, \modified{we utilize PCA to reduce the spectral dimensionality while maintaining the principal components of the HSI data.} This process reduces the number of spectral bands from $C$ to $B$ without altering the spatial dimensions, where $B$ represents the selected spectral bands. The pre-processed data is then represented as $\mathit{I}_{\mathrm{pca}}\in\mathbb{R}^{H\times W\times B}$.

Subsequently, we create 3-D patches from the original HSI data with a window size of $P\times P$. Each 3-D patch, denoted as $\mathit{P}{\in}\mathbb{R}^{P\times P\times B}$, is derived from $\mathit{I}_{\mathrm{pca}}$. The center pixel of each patch is located at $(x_{i},x_{j})$, and the label of this center pixel determines the true label of the patch. Extracting patches around edge pixels presents a challenge. To address this, we apply padding of size $(P-1)/2$ to these edge pixels. Consequently, each patch extends from  $x_{i}-(P-1)/2$ to $x_{i}+(P-1)/2$ in width and from $x_{j}-(P-1)/2$ to $x_{j}+(P-1)/2$ in height, encompassing all spectral bands within this spatial extent. After generating all 3-D patches, we exclude those with labels equal to 0. The remaining patches are then proportionally divided into training and testing datasets.

\subsection{Twin-branch Feature Extraction}

Due to their exceptional capabilities in local feature extraction, CNNs have become widely used in HSI classification. In the proposed method, we utilize two consecutive twin-branch feature extraction (TBFE) blocks to capture both spectral and spatial information from HSI data. Each training sample patch of size $P\times P\times B$ serves as input data for this module. The TBFE module comprises a parallel combination of a 3-D and a 2-D convolution block. Our design aims to seamlessly integrate spectral and spatial information in the early stages of feature extraction by using lightweight structures. The structure of the TBFE block is shown in \modified{Fig.~\ref{fig1}}. Initially, we designed a low-output channel pointwise convolutional layer to adjust the channel dimensionality of HSI. This approach of dimensionality reduction concurrently extracts spectral information while reducing the input channels for subsequent layers, facilitating adjustments in the number of channels for concatenation between successive convolutional layers. Compared to larger convolutional kernels, the use of $\mathit{1\times1}$ \modified{convolution} considerably reduces computational complexity. Following the pointwise convolution, we obtain the feature map $\mathit{F}\in\mathbb{R}^{P\times P\times S}$, where $S$ represents the number of channels obtained after the pointwise convolution. The feature map $\mathit{F}$ is then split into two separate branches, each dedicated to extracting deeper features.

In one branch of the module, a 3-D convolution is integrated with an expand and squeeze mechanism to strengthen the ability to capture spectral features. In particular, for the feature map $\mathit{F}$, we first expand its dimensions by adding an extra channel, transforming it into a 4-D tensor, denoted as $\mathit{F}_{3d}^{\mathrm{in}}\in\mathbb{R}^{P\times P\times S\times 1}$. This transformation prepares the data for 3-D convolution. Padding is applied to ensure that the output dimensions match those of $\mathit{F}_{3d}^{\mathrm{in}}$, resulting in $\mathit{F}_{3d}^{\mathrm{out}}\in\mathbb{R}^{P\times P\times S\times 1}$. Finally, we perform a squeeze operation to convert $\mathit{F}_{3d}^{\mathrm{out}}$ from \modified{the} 4-D space back to the original 3-D space, yielding $\mathit{F}_{3d}\in\mathbb{R}^{P\times P\times S}$, matching the dimensions of $\mathit{F}$. By utilizing convolution with a kernel size of $1\times1\times3$, we fully leverage the automatic learning representation capability of 3-D convolution, enabling the model to concentrate on extracting spectral features more effectively while also minimizing the computational complexity. The computation process of the 3-D convolution block can be expressed as:
\begin{align}
    \mathit{F}_{3d}=\Phi(\mathit{F}\Theta w_{3d}+b_{3d})
\end{align}
where $\mathit{F}_{3d}$ represents the 3-D feature map, $w_{3d}$ and $b_{3d}$ are the weight and bias parameters, respectively. $\Theta$ is the 3-D convolution operator, and $\Phi$ is the activation function.

In the other branch, we utilize 2-D convolution to extract spectral-spatial features. For the input feature map $\mathit{F}$, we employ a $3\times 3$ convolution kernel to filter and capture features across different channels, resulting in $\mathit{F}_{2d}\in\mathbb{R}^{P\times P\times S}$. During this process, padding is applied to the input feature map to preserve information integrity and \modified{ensure} that the output feature map maintains the same dimensions as the input $\mathit{F}$. Following the convolutional layers, activation functions and normalization layers are integrated. This not only introduces non-linearity, enabling the model to learn more complex features but also accelerates training speed and stabilizes the model training process. The computation process of the 2-D convolution block can be expressed as:
\begin{align}
    \mathit{F}_{2d}=\Phi(\mathit{F}\odot w_{2d}+b_{2d})
\end{align}
where $\mathit{F}_{2d}$ represents the 2-D feature map,  $w_{2d}$ and $b_{2d}$ are the weight and bias parameters, respectively. $\odot$ is the 2-D convolution operator, and $\Phi$ is the activation function.

Finally, we merge the feature maps extracted from the two branches to fully leverage the advantages of both convolutions. Specifically, \modified{the feature map $\mathit{F}_{3d}$ from the 3-D convolution branch is concatenated with $\mathit{F}_{2d}$ from the 2-D convolution branch. This fusion strategy combines the strengths of 3-D convolution in capturing spectral dependencies and 2-D convolution in extracting spatial patterns, thereby enhancing the performance of the model to represent spectral-spatial features.} The concatenation process of the two branches can be represented as the following formula:
\begin{align}
    \mathit{M}_{out}=\mathrm{Concat}(\mathit{F}_{3d},\mathit{F}_{2d})
\end{align}
where $Concat(\cdot)$ represents the splicing operation, and $\mathit{M}_{out}\in\mathbb{R}^{P\times P\times 2S}$ is the resulting concatenated feature map.

In addition, we incorporate activation functions and normalization layers at the end of the model to mitigate gradient vanishing or exploding issues and enhance its generalization ability. Furthermore, two 2-D convolutional layers are stacked \modified{to improve} feature extraction.

\subsection{Hybrid Pooling Attention}

\begin{figure}
    \centering
    \includegraphics[width=1\linewidth]{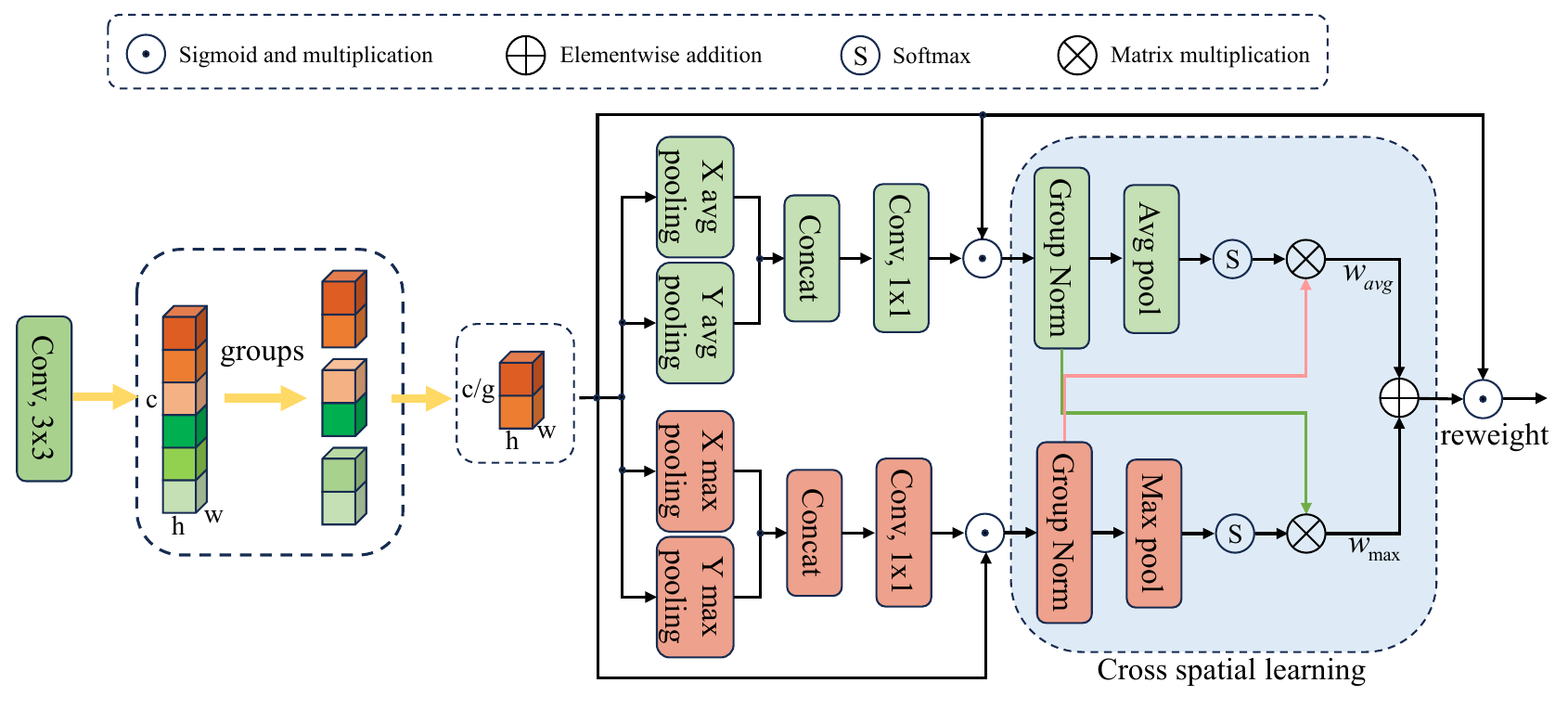}
    \caption{An illustration of the hybrid pooling attention module (HPA). The “g” represents the divided groups. The “X avg Pool” and “X max Pool” denote the 1D horizontal global pooling, as well as the “Y avg Pool” and "X max Pool" denote the 1D vertical global pooling.}
    \label{fig2}
\end{figure}
To fully leverage the spatial information in HSI, an efficient hybrid pooling attention (HPA) module is designed, as shown in Fig.~\ref{fig2}. The HPA module captures effective channel representations without reducing channel dimensionality, \modified{which enhances} pixel-level attention \modified{in} advanced feature maps. It establishes both local and non-local dependencies, thereby improving overall classification performance.

Given the input feature map $\mathit{M}\in\mathbb{R}^{C\times H\times W}$, where $C$ represents the \modified{number} of input channels. $H$ and $W$ denote the spatial size of the input, respectively. The HPA partitions $\mathit{M}$ into $G$ sub-features along the channel dimension to capture different semantic information. Subsequently, the grouped data is reorganized into the batch dimension, reshaping the input data into a format of $C//G\times H\times W$. The resulting grouped data is denoted as $\mathit{M}=\left[M_{0},M_{i},...,M_{G-1}\right]$, where $\mathit{M}_{i}\in\mathbb{R}^{C//G\times H\times W}$. 

In the HPA module, two concurrent paths are utilized to obtain attention-weight representations from the grouped feature maps, with each path focusing on different aspects of the data. Intuitively, average pooling approximates the values within the window, similar to a vague memory of routine experiences. Max pooling captures the peak value within the window, resembling a vivid memory of significant or exceptional events. By combining these two pooling operations, we effectively integrate both general and detailed information, thereby enhancing the representation of the feature maps. In each branch, we perform two 1D global pooling operations along the two spatial dimensions to capture channel attention. Specifically, the average branch employs average pooling, while the max branch utilizes max pooling. In the average pooling branch, we decompose the grouped data into two parallel 1D feature encoding sub-branches to model long-range spatial dependencies. One sub-branch performs average pooling along the height dimension to collect vertical positional information. The 1D global average pooling used to encode global information along the horizontal dimension at position H in channel C can be denoted by:
\begin{align}
    \mathit{Z}_{c-avg}^{H}\left(H\right)=\frac{1}{W}\sum_{0\leq i\leq W}m_{c}\left(H,i\right)
\end{align}
where $m_{c}$ represents the input feature of the c-th channel. This encoding approach enables HPA to capture long-range dependencies along the horizontal direction while maintaining accurate positional details along the vertical direction.

In the other sub-branch, we perform global average pooling along the width dimension, thus it can be regarded as a collection of positional information along the vertical dimension. Therefore, the pooling output at position $W$ in channel $C$ can be represented as:
\begin{align}
    \mathit{Z}_{c-avg}^{W}\left(W\right)=\frac{1}{H}\sum_{0\leq j\leq H}m_{c}\left(j,W\right)
\end{align}

Similarly, in the max pooling branch, we decompose the grouped data into two parallel 1D feature encoding sub-branches. One sub-branch performs max pooling along the horizontal dimension at position $H$ in channel $C$, \modified{and the result} can be formulated as:

\begin{align}
    \mathit{Z}_{c-max}^{H}\left(H\right)=\underset{0\leq i\leq W}{\mathrm{Max}} \; m_{c}\left(H,i\right)
\end{align}
where $m_{c}$ represents the input feature of the $c$-th channel. In the other sub-branch, we perform global max pooling along the width dimension. The pooling output at position $W$ in channel $C$ can be represented as:
\begin{align}
    \mathit{Z}_{c-max}^{W}\left(W\right)=\underset{0\leq j\leq H}{\mathrm{Max}} \; m_{c}\left(j,W\right)
\end{align}

The input features represent global feature information, aiding the model in capturing global information along two spatial directions independently. We concatenate the two encoded features along the height direction of the image \modified{and} then integrate the features from both dimensions using a $1\times1$ convolution. A Sigmoid function is applied to fit the binomial distribution upon linear convolutions. The two-channel attention maps within each group are aggregated through element-wise multiplication, adaptively recalibrating the inter-channel relationships and achieving different cross-channel \modified{interactions} between the two parallel paths in both pooling branches. Given the complementary nature of the precise positional information retained along different spatial directions, recalibrating the raw input features allows HPA to effectively capture and learn intricate low-level feature representations.

Moreover, inspired by its capacity to establish interdependencies among channels and spatial locations, we propose a method for cross-spatial information aggregation across different spatial dimensions. The reweighted feature maps are first processed through a GroupNorm layer to enhance training stability. Subsequently, we separately utilize 2D global average pooling and max pooling to encode global spatial information, obtaining $\mathit{W}_{avg}\in\mathbb{R}^{C\times H\times W}$ and $\mathit{W}_{max}\in\mathbb{R}^{C\times H\times W}$. To achieve efficient computation, the natural non-linear function Softmax is applied to the pooling outputs to fit the linear transformations. By performing matrix dot-product operations between the outputs of the aforementioned \modified{processes} and the matrices obtained after GroupNorm in another branch, we obtain spatial information of different scales collected in the same processing stage. The output feature maps $\mathit{M}^{\prime}$ within each group are calculated by adding the spatial attention weight values generated by the two branches and applying a Sigmoid function. \modified{This can be} expressed as:
\begin{align}
\mathit{M}^{\prime}=\sigma(\mathit{W}_{avg}+\mathit{W}_{max})\odot\mathit{M}
\end{align}
where $\mathit{M}$ represents the input feature map, $\sigma$ represents the Sigmoid function, \modified{and} $\odot$ denotes the elementwise multiplication.

\subsection{Vision Transformer Encoder}

Transformer has achieved state-of-the-art results \modified{in} various vision tasks. As shown in Fig.~\ref{fig1}, the feature maps generated by the preceding HPA module serve as the input to the transformer encoder, which aims to model the interactions among high-level semantic features. Positional information is incorporated through positional embeddings, with each token denoted as $[T_1, T_2, ..., T_N]$. A learnable token $T_{\mathrm{cls}}$ is concatenated with these tokens for the classification task. The positional embeddings $T_{pos}$ are subsequently integrated into the token representations. This process produces a sequence of semantic tokens with embedded positional information, represented as follows:
\begin{align}
\mathit{T}_{\mathrm{in}}=\begin{bmatrix}T_{\mathrm{cls}},T_1,\ldots,T_N\end{bmatrix}+\mathrm{T}_{\mathrm{pos}}.
\end{align}

The core of the transformer is its attention mechanism. It performs linear transformations on the input sequence to obtain Query, Key, and Value representations. The Softmax function is then employed to calculate the weights, resulting in the final attention scores. The self-attention (SA) mechanism can be mathematically represented in the following manner:
\begin{align}
    \mathrm{Attention}(\mathit{Q},\mathit{K},\mathit{V})=\mathrm{softmax}\biggl(\frac{(\mathit{QK})^T}{\sqrt{d_K}}\biggr)\mathit{V}
\end{align}
where $d_K$ denoted the dimensionality of $\mathit{K}$.

Multi-head attention enables the network to focus on different parts of the input sequence simultaneously. By using multiple attention heads, the model can capture diverse relationships and dependencies in the data, leading to a more comprehensive understanding of the input. This mechanism can be represented as follows:
\begin{align}
    \mathrm{MSA}(\mathit{Q},\mathit{K},\mathit{V})=\mathrm{Concat}(\mathrm{SA}_1, \mathrm{SA}_2, \ldots, \mathrm{SA}_h)\mathit{W}
\end{align}
where $h$ denotes the number of heads, $\mathit{W}\in\mathbb{R}^{h\times d_K\times d_w}$ represents the parameter matrix.

Subsequently, the representation matrix obtained in the prior stage is passed through the MLP layer to refine the mapping relationship between spectral features. The MLP typically consists of multiple fully connected layers, each followed by a non-linear activation function. After the MLP layer, LayerNorm is applied to address the vanishing gradient issue and accelerate the training process. The unit with the highest value in the output corresponds to the predicted label of the pixel. 

\subsection{Cross-layer Feature Fusion}

\begin{figure}
    \centering
    \includegraphics[width=0.95\linewidth]{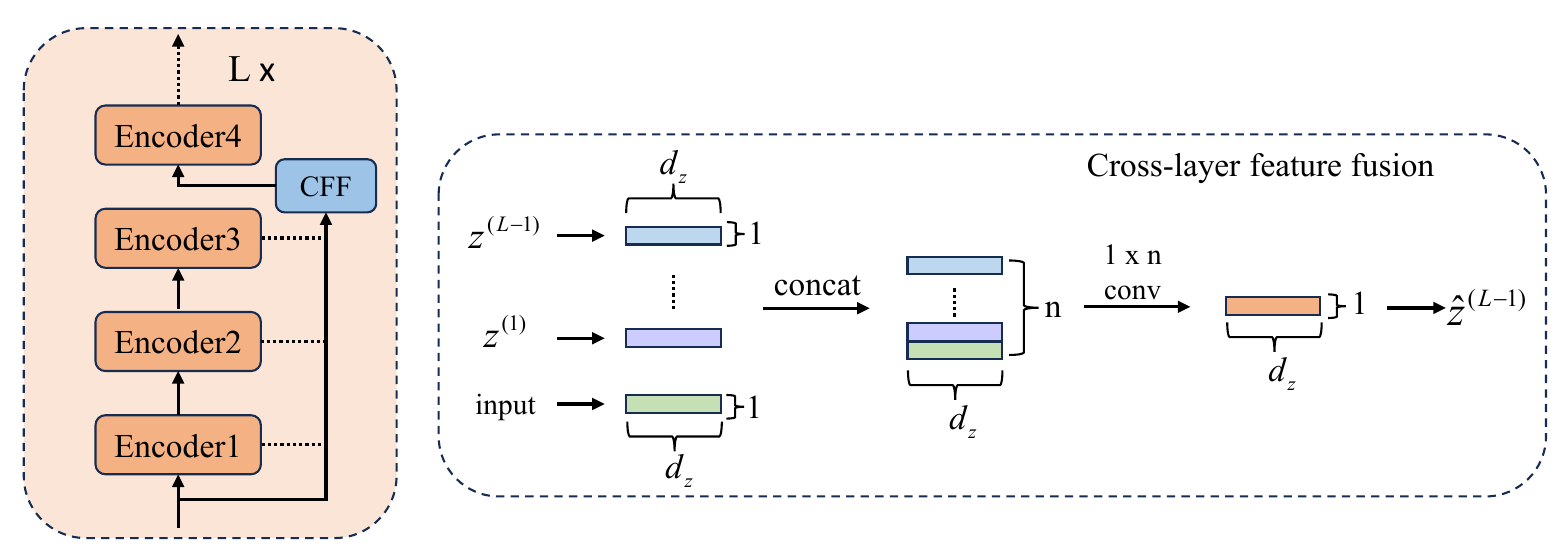}
    \caption{An illustration of the cross-layer feature fusion module (CFF).}
    \label{fig3}
\end{figure}

The skip connection technique has demonstrated its effectiveness as a valuable approach in deep neural networks. To promote better information flow between layers and minimize information loss during the learning process, we designed a cross-layer feature fusion (CFF) module. This module integrates inputs and outputs from different encoders through adaptive cross-layer fusion into the input of the last encoder to generate the final output, as shown in Fig.~\ref{fig3}. Given the input $x_{in}$ and a series of outputs from the corresponding transformer encoder $\mathit{Z}= [z^{(1)},z^{(2)}, ..., z^{(L-1)}]$, the cross-layer feature fusion can be described as:
\begin{align}    \hat{Z}^{(L)}\leftarrow\boldsymbol{w}\bigg[\begin{array}{c}x_{in}\\z^{(1)}\\\vdots\\z^{(L-1)}\end{array}\bigg]
\end{align}
where $z^{(i)}$ represents the output of the $i$-th transformer encoder, $w$ is the learnable network parameter for adaptive fusion, and $\hat{Z}^{(L)}$ denotes the fused \modified{representation} of the input and the outputs from the previous $\mathit{L-1}$ encoders.

\section{Experiments}
\label{sec3}
In this section, we evaluate the proposed method through extensive experiments compared with state-of-the-art methods. We begin with a thorough introduction to the popular datasets used in these experiments. Next, we describe various experimental configurations and perform parameter analyses. We then present and analyze the classification results of our method in comparison with others. Following this, we conduct ablation studies to demonstrate the efficacy of each module. Finally, we analyze the complexity and effectiveness to validate the superiority of our model.

\subsection{Data Description}

\begin{enumerate}[label=\arabic*)]
\item Salinas:  The Salinas dataset was collected by the Airborne Visible/Infrared Imaging Spectrometer (AVIRIS) sensor over the Salinas Valley in the United States. This dataset comprises $512\times 217$ pixels with a \modified{spatial} resolution of 3.7 meters. While it captures ground objects across 224 consecutive spectral bands, typically, 20 water-absorptive bands and noise bands are excluded from the analysis. The dataset contains 16 distinct ground-truth classes.

\begin{table}[]
	\centering
	\caption{\modified{Land-cover classes and the sets of training and testing in the Salinas dataset. \label{tab1}}}
	\resizebox{\linewidth}{!}{
	\begin{tabular}{c|ccc}
		\toprule
		Class No. & Class Name                & Training & Testing \\ \hline
		1         & Brocoli-green-weeds-1     & 10       & 1999    \\
		2         & Brocoli-green-weeds-2     & 19       & 3707    \\
		3         & Fallow                    & 10       & 1966    \\
		4         & Fallow-rough-plow         & 7        & 1387    \\
		5         & Fallow-smooth             & 13       & 2665    \\
		6         & Stubble                   & 20       & 3939    \\
		7         & Celery                    & 18       & 3561    \\
		8         & Grapes-untrained          & 56       & 11215   \\
		9         & Soil-vinyard-develop      & 31       & 6172    \\
		10        & Corn-senesced-green-weeds & 16       & 3262    \\
		11        & Lettuce-romaine-4wk       & 5        & 1063    \\
		12        & Lettuce-romaine-5wk       & 10       & 1917    \\
		13        & Lettuce-romaine-6wk       & 5        & 911     \\
		14        & Lettuce-romaine-7wk       & 5        & 1065    \\
		15        & Vinyard-untrained         & 36       & 7232    \\
		16        & Vinyard-vertical-trellis  & 9        & 1798    \\ \hline
		& Total                     & 270      & 53859   \\ \bottomrule
	\end{tabular}
}
\end{table}

\item  Pavia University: The Pavia University dataset was captured by the Reflective Optics System Imaging Spectrometer (ROSIS) at Pavia University in Northern Italy in 2001. The original image consists of 115 spectral bands ranging from $430$ to $860$ $nm$, with a spatial resolution of $610\times 340$. and a geometric resolution of 1.3 meters per pixel.  Preprocessing was conducted to remove 12 water absorption bands, resulting in 103 bands covering nine distinct ground feature categories.
\begin{table}[]
	\centering
    \caption{\modified{Land-cover classes and the sets of training and testing in the Pavia University dataset. \label{tab2}}}
	\begin{tabular}{c|ccc}
		\toprule
		Class No. & Class Name   & Training & Testing \\ \hline
		1         & Asphalt      & 332      & 6299    \\
		2         & Meadows      & 932      & 17717   \\
		3         & Gravel       & 105      & 1994    \\
		4         & Trees        & 153      & 2911    \\
		5         & Metal Sheets & 67       & 1278    \\
		6         & Bare soil    & 251      & 4778    \\
		7         & Bitumen      & 67       & 1263    \\
		8         & Bricks       & 184      & 3498    \\
		9         & Shadows      & 47       & 900     \\ \hline
		& Total        & 2138     & 40638   \\ \bottomrule
	\end{tabular}
\end{table}

\item  Houston2013: The Houston2013 dataset was acquired using an ITRES CASI-1500 sensor at the University of Houston in Texas in 2012. The image comprises $349\times 1905$ pixels and encompasses 144 spectral bands spanning the wavelength range from 380 to 1050 $nm$, with a spatial resolution of 2.5 meters. All pixels are labeled and classified into 15 land cover classes.

\item \modified{WHU-Hi-HanChuan: The WHU-Hi-HanChuan dataset was captured in 2016 using the Headwall Nano-Hyperspec imaging sensor in Hanchuan, Hubei Province, China. It consists of imagery with a size of $1217 \times 303$ pixels, containing 274 spectral bands ranging from 400 to 1000 $nm$, with a spatial resolution of 0.109 meters. This dataset includes 16 categories for research purposes.}

\item \modified{Houston2018: The Houston2018 dataset was collected at the University of Houston and the surrounding area in 2017. It was used in the 2018 IEEE GRSS Data Fusion Contest. The dataset has a spatial size of  $601 \times 2384$ pixels, with a ground sampling distance of 1 meter. It includes 48 spectral bands ranging from 380 to 1050 $nm$ and represents 20 land cover categories.}

\begin{table}[]
	\centering
	\caption{\modified{Land-cover classes and the sets of training and testing in the Houston2013 dataset. \label{tab3}}}
	\begin{tabular}{c|ccc}
		\toprule
		Class No. & Class Name      & Training & Testing \\ \hline
		1         & Healthy Grass   & 125      & 1126    \\
		2         & Stressed Grass  & 125      & 1129    \\
		3         & Synthetis Grass & 70       & 627     \\
		4         & Tree            & 124      & 1120    \\
		5         & soil            & 124      & 1118    \\
		6         & Water           & 33       & 292     \\
		7         & Residential     & 127      & 1141    \\
		8         & Commercial      & 124      & 1120    \\
		9         & Road            & 125      & 1127    \\
		10        & Highway         & 123      & 1104    \\
		11        & Railway         & 123      & 1112    \\
		12        & Parking Lot 1   & 123      & 1110    \\
		13        & Parking Lot 2   & 47       & 422     \\
		14        & Tennis Court    & 43       & 385     \\
		15        & Running Track   & 66       & 594     \\ \hline
		& Total           & 1502     & 13527   \\ \bottomrule
	\end{tabular}
\end{table}

\begin{table}[]
	\centering
	\caption{\modified{Land-cover classes and the sets of training and testing in the WHU-Hi-HanChuan dataset. \label{tab4}}}
	\begin{tabular}{c|ccc}
		\toprule
		Class No. & Class Name    & Training & Testing \\ \hline
		1         & Strawberry    & 447      & 44288   \\
		2         & Cowpea        & 227      & 22526   \\
		3         & Soybean       & 103      & 10184   \\
		4         & Soyghum       & 54       & 5299    \\
		5         & Water Spinach & 12       & 1188    \\
		6         & Watermelon    & 45       & 4488    \\
		7         & Greens        & 59       & 5844    \\
		8         & Trees         & 180      & 17798   \\
		9         & Grass         & 95       & 9374    \\
		10        & Red Roof      & 105      & 10411   \\
		11        & Gray Roof     & 169      & 16742   \\
		12        & Plastic       & 37       & 3642    \\
		13        & Bare Soil     & 91       & 9025    \\
		14        & Road          & 186      & 18374   \\
		15        & Bright Object & 11       & 1125    \\
		16        & Water         & 754      & 74647   \\ \hline
		& Total         & 2575     & 254955  \\ \bottomrule
	\end{tabular}
\end{table}

\end{enumerate}

\modified{Tables~\ref{tab1}-\ref{tab5} list} the land-cover category names, along with the numbers of training and testing samples utilized in the experiments for the \modified{five} aforementioned datasets.

\subsection{Experimental Configuration}

\subsubsection{Evaluation Metrics}
To evaluate the performance of the proposed method compared to other methods, we conducted a quantitative evaluation using three widely accepted metrics: overall accuracy (OA), average accuracy (AA), and Kappa coefficient ($\kappa$). The OA assesses the effectiveness of the classification by calculating the proportion of accurately predicted samples out of the total samples in all categories combined. The AA represents the average accuracy of individual classes and reflects the method's performance across different categories. The $\kappa$ coefficient evaluates the level of agreement between the model's predictions and the ground truth, providing a comprehensive measure of performance. In addition, a higher value for each metric indicates better classification performance.

\begin{table}[]
	\caption{\modified{Land-cover classes and the sets of training and testing in the Houston2018 dataset. \label{tab5}}}
	\begin{tabular}{c|ccc}
		\toprule
		Class No. & Class Name                & Training & Testing \\ \hline
		1         & Healthy grass             & 980      & 8819    \\
		2         & Stressed grass            & 3250     & 29252   \\
		3         & Artificial turf           & 68       & 616     \\
		4         & Evergreen trees           & 1311     & 11796   \\
		5         & Deciduous trees           & 481      & 4329    \\
		6         & Bare earth                & 451      & 4064    \\
		7         & Water                     & 26       & 239     \\
		8         & Residential buildings     & 3827     & 34441   \\
		9         & Non-residential buildings & 22114    & 199031  \\
		10        & Roads                     & 4118     & 37060   \\
		11        & Sidewalks                 & 2861     & 25748   \\
		12        & Crosswalks                & 140      & 1259    \\
		13        & Major thoroughfares       & 4493     & 40440   \\
		14        & Highways                  & 951      & 8556    \\
		15        & Railways                  & 694      & 6243    \\
		16        & Paved parking lots        & 1072     & 9653    \\
		17        & Unpaved parking lots      & 13       & 116     \\
		18        & Cars                      & 483      & 4352    \\
		19        & Trains                    & 462      & 4160    \\
		20        & Stadium seats             & 682      & 6142    \\ \bottomrule
		& Total                     & 48479    & 436316  \\ \hline
	\end{tabular}
\end{table}

\subsubsection{Experimental Environment}
The verification experiments for the proposed method were conducted in a PyTorch environment. Training and testing were performed on a server equipped with an Intel(R) Xeon(R) Gold 6226R CPU @ 2.90GHz and an NVIDIA GeForce RTX 3090 24GB GPU. We selected the Adam optimizer with an initial learning rate of 1e-3. The batch size was set to 64, and the training process was carried out for 100 epochs.

\subsection{Parameter Analysis}

\begin{figure*}
    \centering
    \includegraphics[width=\linewidth]{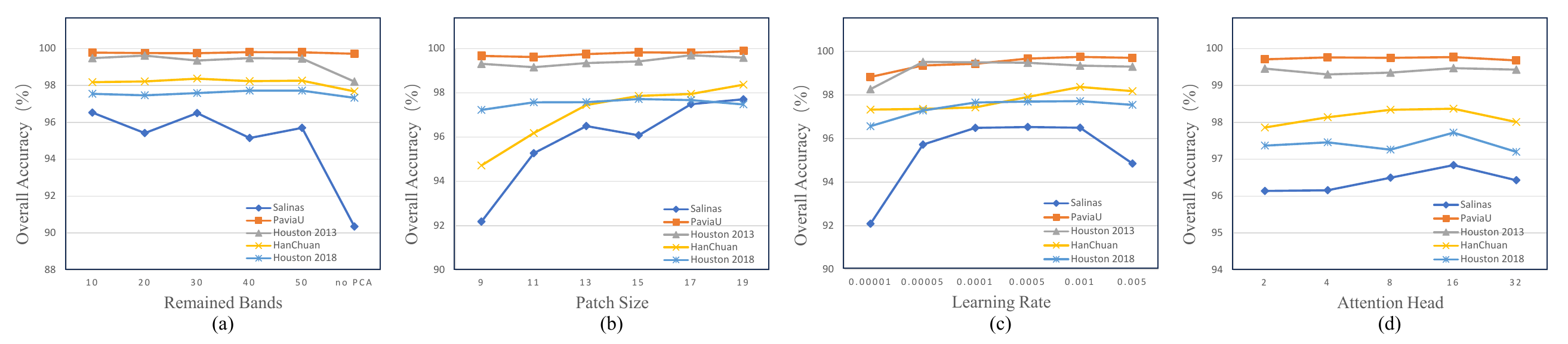}
    \caption{\modified{The OA impact curves of different parameters. (a) Remained bands. (b) Patch size. (c) Learning rate. (d) Attention Head.}}
    \label{fig4}
\end{figure*}

\begin{table*}[]
	\centering
	\caption{\modified{Quantitative results of experiments with different number of encoders on five datasets. The best results are highlighted in bold.\label{tab6}}}
	\resizebox{\linewidth}{!}{
	\begin{tabular}{c|cccccccccc|c}
		\toprule
		\multirow{2}{*}{\begin{tabular}[c]{@{}c@{}}Number of \\ Encoders\end{tabular}} & \multicolumn{2}{c}{Salinas}     & \multicolumn{2}{c}{Pavia University} & \multicolumn{2}{c}{Houston2013} & \multicolumn{2}{c}{WHU-Hi-HanChuan} & \multicolumn{2}{c|}{Houston2018} & \multirow{2}{*}{\begin{tabular}[c]{@{}c@{}}Number of\\ Parameters\end{tabular}} \\ \cline{2-11}
		& OA(\%)         & $\kappa\times100$          & OA(\%)            & $\kappa\times100$            & OA(\%)         & $\kappa\times100$         & OA(\%)           & $\kappa\times100$            & OA(\%)          & $\kappa\times100$          &                                                                                 \\ \hline
		1                                                                              & 97.87          & 97.63          & 99.43             & 99.25            & 99.34          & 99.30          & 98.18            & 97.87            & 97.59           & 96.83          & \textbf{113.38k}                                                                \\
        2                                                                              & 98.52          & 98.35          & 99.49             & 99.32            & 99.49          & 99.45          & 98.29            & 98.01            & 97.62           & 96.86          & 131.43k                                                                         \\
        3                                                                              & \textbf{98.67} & \textbf{98.52} & 99.46             & 99.28            & 99.46          & 99.39          & 98.33            & 98.06            & 97.57           & 96.79          & 149.47k                                                                         \\
        4                                                                              & 98.61          & 98.45          & \textbf{99.92}    & \textbf{99.89}   & \textbf{99.65} & \textbf{99.62} & \textbf{98.37}   & \textbf{98.09}   & \textbf{97.72}  & \textbf{96.99} & 167.49K                                                                         \\
        5                                                                              & 98.59          & 98.37          & 99.34             & 99.12            & 99.40          & 99.35          & 98.24            & 97.98            & 97.65           & 96.87          & 185.51k                                                                         \\ \bottomrule
	\end{tabular}
}
\end{table*}

We examined the influence of various crucial parameters on the classification results of the proposed network. These parameters include reduced spectral dimensions, patch size, network learning rate, and the number of attention heads. \modified{Additionally, we analyzed the impact of different numbers of encoders on five datasets, as shown in Fig.~\ref{fig4} and Table~\ref{tab6}.}

Remained Spectral Dimensions: To assess the effect of varying spectral dimensions on classification performance while holding all other parameters fixed, we evaluated candidate values of 10, 20, 30, and 40. Furthermore, the classification results without PCA-based dimensionality reduction were assessed. Fig.~\ref{fig4}(a) presents the OA of the model across different spectral dimensions for the \modified{five} datasets. The comparison chart clearly indicates that the model performs worst across all datasets when no dimensionality reduction is applied. \modified{After performing feature reduction, the Houston2013 dataset exhibits an improvement in OA as the spectral dimensions increase, followed by a gradual decline. For the Salinas dataset, the OA peaks at dimensions 10 and 30, achieving a 6.18\% increase compared to the results without PCA. In the case of the Hanchuan dataset, which contains the largest number of spectral bands, classification performance is the lowest when only 10 bands are retained and improves steadily as more spectral dimensions are included. Conversely, the PU and Houston2018 datasets demonstrate stable performance initially, followed by a gradual upward trend. Taking into account both classification accuracy and computational efficiency, we selected 30 as the optimal number of retained spectral dimensions.}

Patch Size: To investigate how patch size influences classification performance, we fixed all other parameters and evaluated candidate values of 9, 11, 13, 15, 17 and 19 for the patch size. Fig.~\ref{fig4}(b) presents the OA of the model across different patch sizes for the \modified{five} datasets. \modified{It is evident that the OA of the Houston2013 dataset achieves optimal results with a patch size of 17, while the PU dataset remains relatively stable. For the Hanchuan dataset, OA increases steadily as the patch size grows. In contrast, the OA for the Houston2018 dataset initially rises and then declines. For the SA dataset, the OA decreased at a patch size of 15 before increasing steadily. Consequently, we selected a patch size of 17 for the Houston2013 dataset, 15 for the Houston2018 dataset and 19 for the remaining three datasets.}

Learning Rate: The learning rate is a crucial factor in the model training process, significantly affecting the speed of convergence. We evaluated a range of candidate values: $5e-5$, $1e-4$, $5e-4$, $1e-3$, and $5e-3$. Fig.~\ref{fig4}(c) illustrates the effect of various learning rates on OA. It is evident that, across the \modified{five} datasets, the OA of the proposed network initially increases and then decreases, indicating an optimal learning rate. \modified{Moreover, within the range of $[1e-4, 1e-3]$, the OA for the PU and Hanchuan datasets exhibits continuous improvement, while the other three datasets remain relatively stable, demonstrating robustness to changes in the learning rate.} Based on these observations, we selected $1e-3$ as the learning rate for subsequent classifications.

Number of attention heads: The number of attention heads is another crucial parameter that influences classification. Keeping other parameters constant, we evaluated a range of candidate values: 2, 4, 8, 16, and 32. Fig.~\ref{fig4}(d) demonstrates the effect of different numbers of attention heads on OA. \modified{When the number of attention heads is within the range of [4, 16], the OA value of the Houston2018 dataset exhibits a trend of initially decreasing and then increasing, while the PU dataset remains stable, and the other three datasets show a consistent and steady improvement. This indicates that increasing the number of attention heads helps explore targets from multiple perspectives. when the number of heads increases further, excessive redundant information is introduced, leading to a decline in performance.} We selected 16 attention heads, as this configuration achieved optimal network performance.

\modified{Number of encoders: To evaluate the impact of the number of encoders on classification performance, we conducted experiments with different numbers of encoders, ranging from 1 to 5, while keeping other parameters constant. Table~\ref{tab6} summarizes the quantitative results across five datasets. As shown in the table, increasing the number of encoders led to varying degrees of improvement in both OA and Kappa across different datasets. Due to the smaller spatial dimension of the Salinas dataset, the highest OA and Kappa values were achieved with 3 encoders. For the remaining datasets, which have relatively larger spatial dimensions, more encoders were needed to capture the feature representations, thus the best classification performance was achieved when the number of encoders was set to 4. However, further increasing the number of encoders resulted in parameter count and the introduction of redundant information, which caused a decline in performance. Based on these observations, we selected 3 encoders for the Salinas dataset and 4 encoders for the remaining datasets.}

\subsection{Classification Results}

To validate the effectiveness of our proposed classification approach, we conducted comparative experiments with several well-established methods, including SVM\cite{1323134}, 2D-CNN\cite{7450160}, 3D-CNN\cite{chen2016deep}, HybridSN\cite{8736016}, SpectralFormer\cite{9627165}, SSFTT\cite{9684381}, \modified{MASSFormer \cite{10506482}, SS-Mamba\cite{rs16132449}}, and our proposed method. To ensure a fair comparison, we evaluated the classification performance of these methods based on randomly selected samples from each dataset: 0.5\% per category for the Salinas dataset, 5\% per category for the Pavia University dataset, 10\% per category for the Houston2013 dataset, \modified{1\% per category for the WHU-Hi-HanChuan dataset, and 10\% per category for the Houston2018 dataset, as detailed in Tables~\ref{tab1}-\ref{tab5}. }

\begin{table*}[]
	\centering
	\caption{Classification results of different methods on the Salina dataset with 0.5\% training samples\label{tab7}}
	\resizebox{\linewidth}{!}{
		\begin{tabular}{c|ccccccccc}
			\toprule
			No.              & SVM\cite{1323134} & 2D-CNN\cite{7450160}    & 3D-CNN\cite{chen2016deep}     & HybridSN\cite{8736016} & SpectralFormer\cite{9627165} & SSFTT\cite{9684381}           & \modified{MASSFormer\cite{10506482}}      & \modified{SS-Mamba\cite{rs16132449}}        & Ours            \\ \hline
			1                 & 48.48 & 99.88  & 96.13  & 99.62           & 93.14          & 99.97           & \textbf{100.00} & \textbf{100.00} & 99.10           \\
			2                 & 61.53 & 96.30  & 99.69  & 99.99           & 98.75          & 99.88           & \textbf{100.00} & \textbf{100.00} & \textbf{100.00} \\
			3                 & 78.39 & 89.73  & 88.35  & 98.75           & 93.57          & \textbf{100.00} & \textbf{100.00} & 99.94           & \textbf{100.00} \\
			4                 & 61.54 & 88.31  & 97.31  & 81.12           & 90.39          & 99.24           & 96.87           & 99.83           & \textbf{99.93}  \\
			5                 & 94.04 & 97.88  & 94.65  & 98.62           & 81.20          & 98.54           & \textbf{100.00} & 98.25           & 99.10           \\
			6                 & 87.76 & 73.46  & 99.69  & 98.96           & 99.90          & 99.85           & 98.95           & \textbf{99.99}  & 99.47           \\
			7                 & 86.21 & 89.63  & 99.41  & 98.26           & 98.69          & 99.53           & 99.89           & \textbf{99.97}  & 98.99           \\
			8                 & 91.12 & 80.96  & 75.61  & 95.81           & 80.56          & 93.31           & 97.90           & 90.58           & \textbf{98.22}  \\
			9                 & 98.33 & 70.98  & 95.96  & \textbf{100.00} & 94.67          & 99.55           & \textbf{100.00} & \textbf{100.00} & \textbf{100.00} \\
			10                & 61.52 & 84.65  & 89.81  & 98.87           & 88.43          & 98.54           & \textbf{99.76}  & 99.49           & 98.28           \\
			11                & 67.00 & 90.96  & 62.87  & 87.69           & 75.72          & \textbf{100.00} & 99.70           & 99.72           & 98.40           \\
			12                & 59.61 & 91.46  & 99.06  & 98.00           & 97.37          & 99.29           & 99.92           & \textbf{100.00} & \textbf{100.00} \\
			13                & 96.45 & 85.65  & 99.65  & 85.18           & 96.48          & 96.31           & 88.34           & \textbf{99.92}  & 92.65           \\
			14                & 88.61 & 73.71  & 96.22  & 95.04           & 92.71          & 96.82           & 93.62           & \textbf{99.88}  & 99.06           \\
			15                & 68.40 & 87.52  & 69.73  & 94.80           & 77.63          & 89.82           & 91.59           & 95.01           & \textbf{96.74}  \\
			16                & 98.72 & 96.16  & 74.48  & 97.79           & 74.77          & 99.47           & \textbf{99.99}  & 99.80           & 98.39           \\ \hline
			OA(\%)            & 73.26 & 85.32  & 87.09  & 96.82           & 88.17          & 96.78           & 97.92           & 97.21           & \textbf{98.67}  \\
			AA(\%)            & 67.15 & 87.40  & 89.91  & 95.53           & 89.62          & 98.13           & 97.91           & \textbf{98.90}  & 98.64           \\
			$\kappa\times100$ & 69.16 & 83.97  & 85.64  & 96.46           & 86.85          & 96.41           & 97.68           & 96.90           & \textbf{98.52}  \\ \bottomrule
		\end{tabular}
	}
\end{table*}

\begin{figure*}
	\centering
	\includegraphics[width=\linewidth]{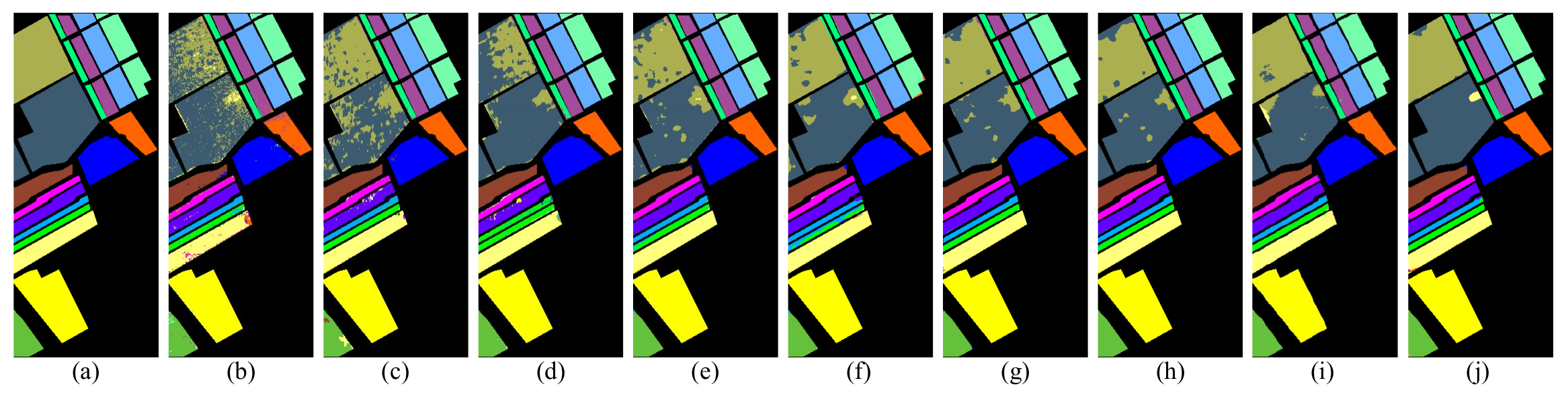}
	\caption{\modified{Classification maps of the Salinas dataset. (a) Ground-truth map, (b) SVM, (c) 2-D-CNN, (d) 3-D-CNN, (e) Hybrid, (f)SpectralFormer, (g)SSFTT, (h)MASSFormer, (i)SS-Mamba, (j)Ours.\label{fig5}}}
\end{figure*}

\begin{figure*}
	\centering
	\includegraphics[width=\linewidth]{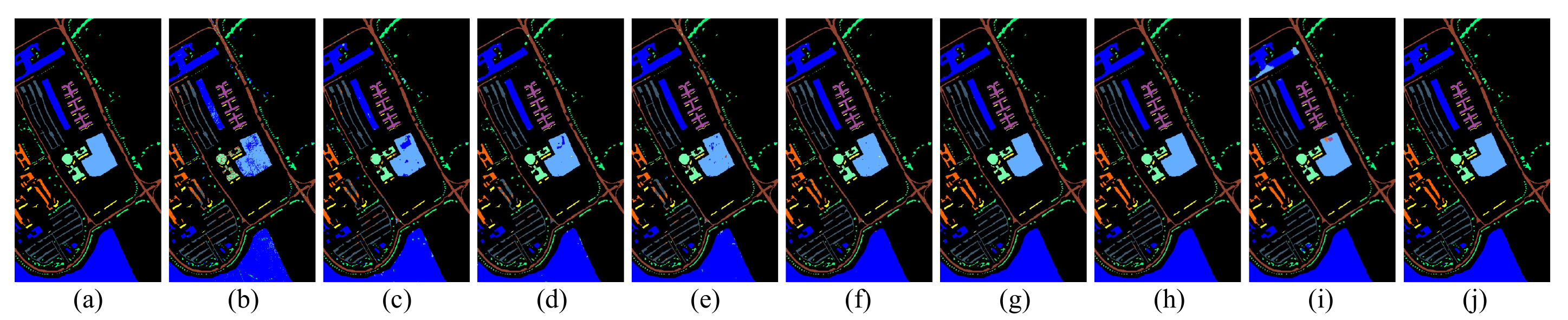}
	\caption{\modified{Classification maps of the Pavia Universit dataset. (a) Ground-truth map, (b) SVM, (c) 2-D-CNN, (d) 3-D-CNN, (e) Hybrid, (f)SpectralFormer, (g)SSFTT, (h)MASSFormer, (i)SS-Mamba, (j)Ours.\label{fig6}}}
\end{figure*}

\begin{figure*}
	\centering
	\includegraphics[width=\linewidth]{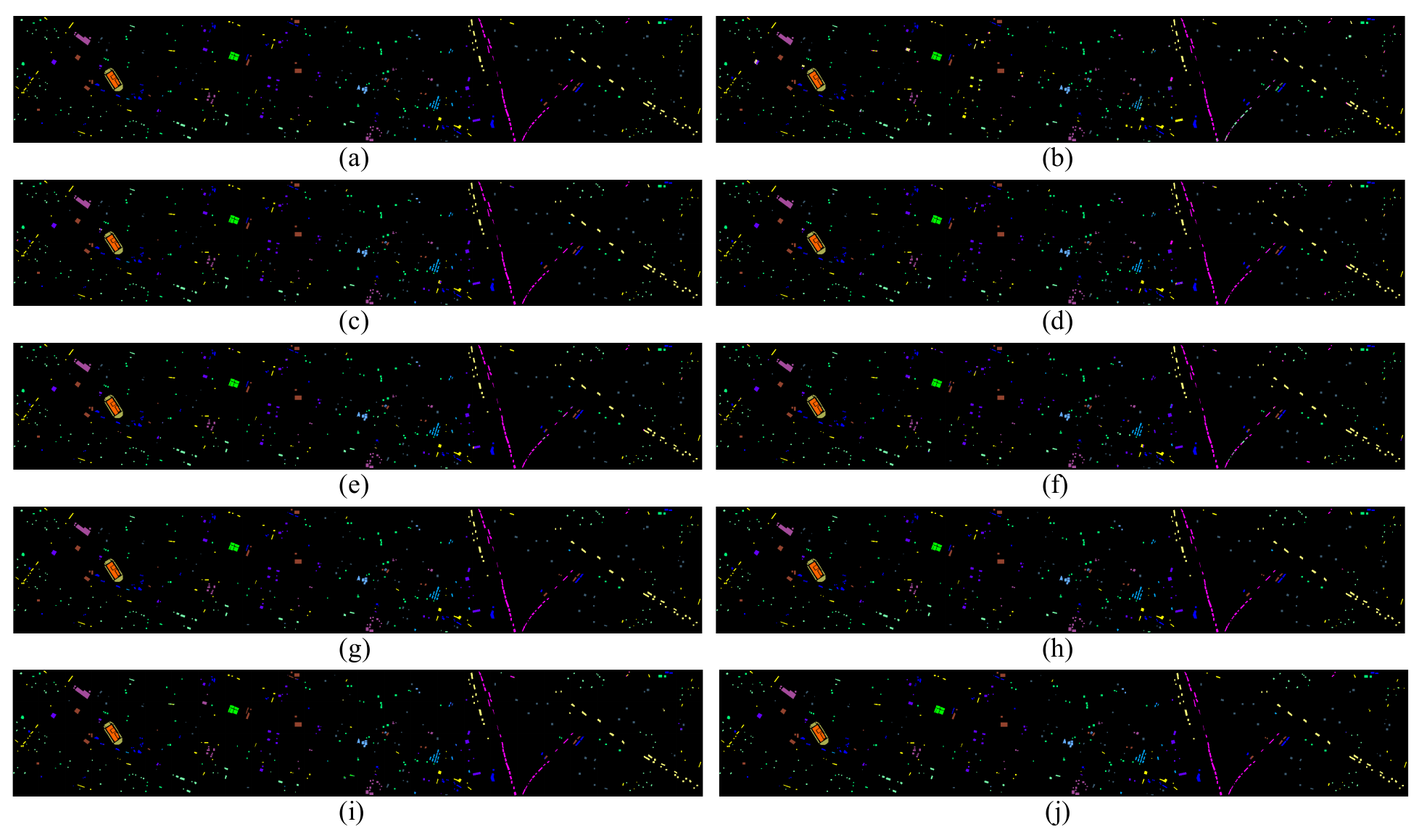}
	\caption{\modified{Classification maps of the Houston2013 dataset. (a) Ground-truth map, (b) SVM, (c) 2-D-CNN, (d) 3-D-CNN, (e) Hybrid, (f)SpectralFormer, (g)SSFTT, (h)MASSFormer, (i)SS-Mamba, (j)Ours.\label{fig7}}}
\end{figure*}

\begin{figure*}
	\centering
	\includegraphics[width=\linewidth]{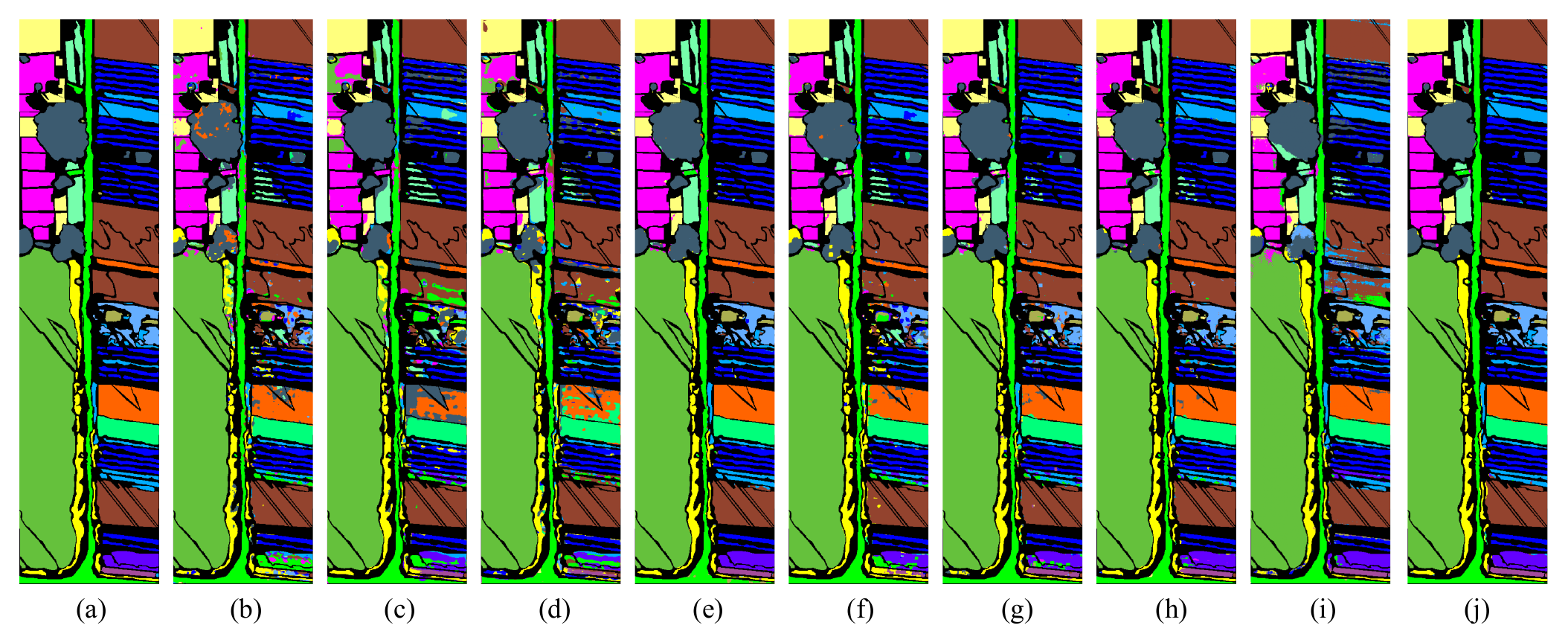}
	\caption{\modified{Classification maps of the WHU-Hi-HanChuan dataset. (a) Ground-truth map, (b) SVM, (c) 2-D-CNN, (d) 3-D-CNN, (e) Hybrid, (f)SpectralFormer, (g)SSFTT, (h)MASSFormer, (i)SS-Mamba, (j)Ours.\label{fig8}}}
\end{figure*}

\begin{figure*}
	\centering
	\includegraphics[width=\linewidth]{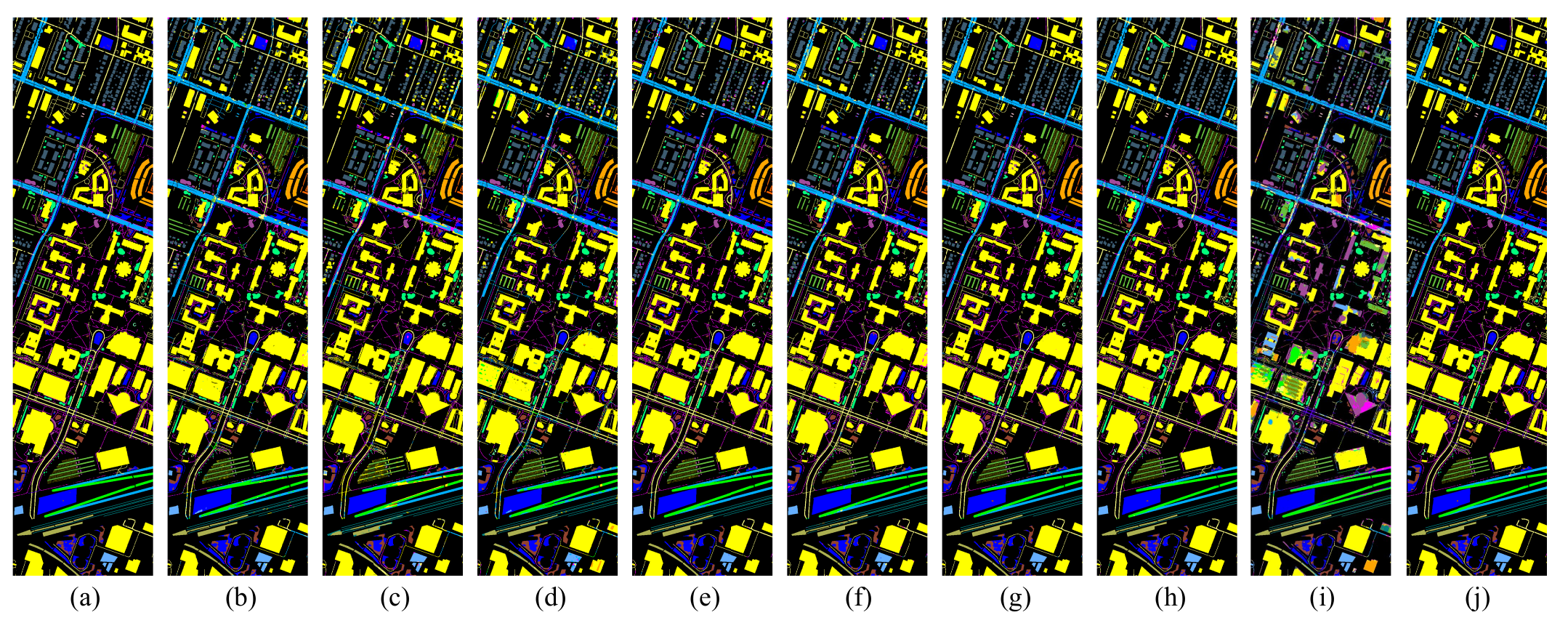}
	\caption{\modified{Classification maps of the Houston2018 dataset. (a) Ground-truth map, (b) SVM, (c) 2-D-CNN, (d) 3-D-CNN, (e) Hybrid, (f)SpectralFormer, (g)SSFTT, (h)MASSFormer, (i)SS-Mamba, (j)Ours.\label{fig9}}}
\end{figure*}

Quantitative Results and Analysis: Tables~\ref{tab7}-\ref{tab11} present the single-class accuracy, OA, AA, and $\kappa \times 100$ for our proposed method and the comparison methods across the \modified{five} datasets, with the best results highlighted in bold. \modified{Additionally, to intuitively visualize the classification performance of different models, Figs.~\ref{fig5}–\ref{fig9} illustrate the classification result maps generated by our proposed method alongside those of the comparison methods.}

\modified{1) Salinas: The classification results for the Salinas dataset are shown in Table~\ref{tab7}. Due to the lack of feature extraction, SVM exhibits the worst classification performance. Transformer-based methods perform well by modeling the global dependencies within the input sequences. Our proposed method achieves the best classification performance in terms of OA and $\kappa \times 100$, with values of 98.67\% and 98.52\%, respectively. Thanks to effective feature extraction, we improve the OA by 0.75\% compared to the second-place method, MASSFormer. However, due to class imbalance, the classification accuracy for the Lettuce-romaine-6wk category is relatively low. This is because random sampling, relying on percentage-based sampling, results in fewer training samples for this category. As a result, our AA value is 0.26\% lower than that of SS-Mamba. Overall, our OA and $\kappa \times 100$ values are 1.46\% and 1.62\% higher than those of SS-Mamba, respectively. Furthermore, the classification result maps shown in Figs.~\ref{fig5} clearly demonstrate that our method produces the most accurate and smooth classification maps, especially in the Fallow and Fallow-rough-plowf categories.}

\modified{2) Pavia University: Table~\ref{tab8} clearly shows that for the Pavia University dataset, known for its high spatial complexity, HybridSN combines 3D and 2D convolutions to extract richer local texture information from HSI blocks but struggles to capture global features. In contrast, our method effectively distinguishes subtle differences between land cover classes. In 4 out of 9 categories, our method achieves 100.00\% classification accuracy and outperforms all other methods in OA, AA, and $\kappa \times 100$. The corresponding classification maps for different models are shown in Figs.~\ref{fig6}, where our method exhibits significantly higher fragmentation in the classification results, further demonstrating its superior performance.}

\modified{3) Houston2013: The classification results for the Houston2013 dataset are shown in Table~\ref{tab9}. The SSFTT and MASSFormer methods combine the advantages of CNN and Transformer, leading to good classification performance. Despite the increased number of spectral bands, our method still effectively learns key spectral information and extracts more discriminative features. Specifically, our method achieves the highest values for OA, AA, and $\kappa \times 100$, with 99.65\%, 99.52\%, and 99.62\%, respectively, surpassing all convolution-based, Transformer-based, and Mamba-based comparison methods. Additionally, even with fewer classes in the Running Track category, our method achieves the best classification results. The classification maps shown in Figs.~\ref{fig7} demonstrate that our method produces more homogeneous classification results.}

\begin{table*}[]
	\centering
	\caption{Classification results of different methods on the Pavia University dataset with 5\% training samples\label{tab8}}
	\resizebox{\linewidth}{!}{
	\begin{tabular}{c|ccccccccc}
		\toprule
		No.              & SVM\cite{1323134} & 2D-CNN\cite{7450160}    & 3D-CNN\cite{chen2016deep}     & HybridSN\cite{8736016} & SpectralFormer\cite{9627165} & SSFTT\cite{9684381}           & \modified{MASSFormer\cite{10506482}}      & \modified{SS-Mamba\cite{rs16132449}}        & Ours            \\ \hline
		1                  & 91.22 & 93.28  & 93.30  & 96.51           & 96.90          & \textbf{99.94} & 99.37           & 95.81           & 99.87           \\
		2                  & 97.78 & 94.90  & 94.21  & 94.49           & 95.59          & 99.97          & 99.99           & 92.54           & \textbf{100.00} \\
		3                  & 34.95 & 75.55  & 90.19  & 94.18           & 78.90          & 98.97          & 96.91           & 99.70           & \textbf{99.85}  \\
		4                  & 81.55 & 93.87  & 91.29  & 99.55           & 86.31          & 98.81          & 98.76           & 99.14           & \textbf{99.69}  \\
		5                  & 98.69 & 97.88  & 95.47  & 96.71           & 93.91          & 99.80          & \textbf{100.00} & \textbf{100.00} & \textbf{100.00} \\
		6                  & 41.50 & 70.05  & 93.86  & 99.43           & 82.54          & 99.96          & 99.85           & 99.00           & \textbf{100.00} \\
		7                  & 32.17 & 70.92  & 81.84  & \textbf{100.00} & 88.90          & 99.92          & 99.92           & 99.98           & \textbf{100.00} \\
		8                  & 89.74 & 90.96  & 80.36  & 95.97           & 95.17          & 99.14          & 98.52           & \textbf{99.72}  & 99.69           \\
		9                  & 96.87 & 90.78  & 92.58  & 95.55           & 93.97          & 98.05          & 98.66           & \textbf{99.99}  & 99.89           \\ \hline
		OA(\%)             & 82.76 & 90.19  & 92.01  & 98.16           & 93.17          & 99.70          & 99.48           & 95.78           & \textbf{99.92}  \\
		AA(\%)             & 71.44 & 87.31  & 90.45  & 97.35           & 90.27          & 99.40          & 99.06           & 98.31           & \textbf{99.88}  \\
		$\kappa \times100$ & 76.28 & 87.52  & 90.87  & 97.57           & 91.23          & 99.61          & 99.31           & 94.51           & \textbf{99.89}  \\ \bottomrule
	\end{tabular}
}
\end{table*}

\begin{table*}[]
	\centering
	\caption{Classification results of different methods on the Houston 2013 dataset with 10\% training samples\label{tab9}}
	\resizebox{\linewidth}{!}{
	\begin{tabular}{c|ccccccccc}
		\toprule
		No.              & SVM\cite{1323134} & 2D-CNN\cite{7450160}    & 3D-CNN\cite{chen2016deep}     & HybridSN\cite{8736016} & SpectralFormer\cite{9627165} & SSFTT\cite{9684381}           & \modified{MASSFormer\cite{10506482}}      & \modified{SS-Mamba\cite{rs16132449}}        & Ours            \\ \hline
		1                 & 93.65 & 94.02  & 93.72           & 98.85           & 98.68           & 99.47           & \textbf{99.56}  & 93.47           & 99.29           \\
		2                 & 97.29 & 96.30  & 94.91           & 99.73           & 98.97           & 99.83           & 99.65           & 97.45           & \textbf{99.91}  \\
		3                 & 96.87 & 89.73  & 95.54           & 99.84           & 99.49           & 99.04           & 99.68           & \textbf{100.00} & 99.84           \\
		4                 & 93.66 & 98.31  & 94.91           & 96.07           & 99.72           & 99.29           & \textbf{100.00} & 99.41           & \textbf{100.00} \\
		5                 & 98.30 & 97.88  & \textbf{100.00} & 99.47           & 99.72           & \textbf{100.00} & \textbf{100.00} & 98.22           & \textbf{100.00} \\
		6                 & 84.25 & 73.46  & 90.06           & \textbf{100.00} & 96.38           & \textbf{100.00} & 99.32           & 94.49           & 99.32           \\
		7                 & 82.65 & 89.62  & 91.84           & 97.63           & 96.01           & 97.72           & 97.75           & 94.46           & \textbf{98.69}  \\
		8                 & 56.61 & 80.96  & 80.36           & 97.65           & 96.59           & 97.86           & 99.82           & 81.10           & \textbf{100.00} \\
		9                 & 73.91 & 70.98  & 93.58           & 98.67           & 94.17           & 99.67           & 99.30           & 93.17           & \textbf{100.00} \\
		10                & 83.51 & 84.65  & 94.28           & 99.00           & 98.95           & \textbf{100.00} & \textbf{100.00} & 99.75           & \textbf{100.00} \\
		11                & 68.97 & 90.96  & 92.64           & 99.28           & 97.81           & 99.28           & 99.55           & 96.03           & \textbf{100.00} \\
		12                & 60.72 & 91.46  & 94.37           & 99.46           & 97.04           & 99.32           & 99.28           & 90.64           & \textbf{99.64}  \\
		13                & 27.69 & 85.65  & 90.97           & 98.10           & 85.96           & 98.82           & \textbf{100.00} & 91.25           & 96.45           \\
		14                & 93.25 & 73.71  & 94.58           & \textbf{100.00} & \textbf{100.00} & \textbf{100.00} & 99.23           & \textbf{100.00} & 99.74           \\
		15                & 99.66 & 99.52  & 99.65           & 99.49           & 99.96           & 99.69           & \textbf{100.00} & \textbf{100.00} & \textbf{100.00} \\ \hline
		OA(\%)            & 80.92 & 89.05  & 93.73           & 98.80           & 97.40           & 99.41           & 99.52           & 94.93           & \textbf{99.65}  \\
		AA(\%)            & 79.61 & 87.82  & 93.56           & 98.93           & 97.10           & 99.45           & 99.51           & 95.30           & \textbf{99.52}  \\
		$\kappa\times100$ & 79.33 & 88.15  & 93.63           & 98.71           & 97.21           & 99.36           & 99.48           & 94.52           & \textbf{99.62}  \\ \bottomrule
	\end{tabular}
}
\end{table*}

\begin{table*}[]
     \centering
     \caption{\modified{Classification results of different methods on the WHU-Hi-HanChuan dataset with 1\% training samples\label{tab10}}}
     \resizebox{\linewidth}{!}{
	\begin{tabular}{c|ccccccccc}
		\toprule
		No.              & SVM\cite{1323134} & 2D-CNN\cite{7450160}    & 3D-CNN\cite{chen2016deep}     & HybridSN\cite{8736016} & SpectralFormer\cite{9627165} & SSFTT\cite{9684381}           & MASSFormer\cite{10506482}      & SS-Mamba\cite{rs16132449}        & Ours            \\ \hline
		1                 & 95.95 & 95.53          & 95.19  & 99.26          & 94.40          & 98.49 & 99.01      & 92.13          & \textbf{99.56} \\
		2                 & 86.68 & 77.54          & 87.84  & 96.36          & 83.17          & 94.06 & 96.50      & 86.50          & \textbf{98.22} \\
		3                 & 68.12 & 47.44          & 84.48  & 97.16          & 94.51          & 93.57 & 97.38      & 91.85          & \textbf{98.67} \\
		4                 & 87.04 & 94.92          & 91.21  & \textbf{99.67} & 95.66          & 98.84 & 95.19      & 99.09          & 98.87          \\
		5                 & 53.11 & 72.05          & 52.01  & 94.71          & 67.98          & 89.52 & 90.66      & 99.72          & \textbf{99.83} \\
		6                 & 34.63 & 15.51          & 24.85  & 70.55          & 42.64          & 73.08 & 83.91      & 85.59          & \textbf{91.69} \\
		7                 & 91.58 & 79.96          & 84.39  & 93.35          & 79.41          & 92.63 & 92.09      & \textbf{96.37} & 95.11          \\
		8                 & 69.83 & 90.40          & 77.17  & 93.32          & 80.72          & 94.17 & 91.84      & 82.17          & \textbf{97.62} \\
		9                 & 70.55 & 75.11          & 73.30  & 90.70          & 71.28          & 95.42 & 93.01      & 88.12          & \textbf{96.75} \\
		10                & 90.47 & 92.93          & 90.42  & 97.94          & 95.18          & 98.14 & 98.69      & 97.21          & \textbf{99.16} \\
		11                & 83.00 & 79.85          & 93.98  & 96.53          & 93.47          & 97.85 & 98.44      & 93.56          & \textbf{98.74} \\
		12                & 42.94 & 58.07          & 30.43  & 85.85          & 46.84          & 87.41 & 93.51      & \textbf{98.73} & 97.09          \\
		13                & 67.81 & 54.12          & 62.41  & 81.25          & 70.87          & 79.77 & 80.65      & 82.00          & \textbf{89.41} \\
		14                & 89.52 & 89.40          & 85.93  & 94.31          & 86.27          & 95.30 & 96.03      & 87.42          & \textbf{98.07} \\
		15                & 66.04 & 23.12          & 79.31  & 41.33          & 64.25          & 80.13 & 82.87      & \textbf{92.39} & 79.73          \\
		16                & 98.31 & \textbf{99.96} & 99.36  & 99.83          & 99.29          & 99.48 & 99.79      & 96.51          & \textbf{99.96} \\ \hline
		OA(\%)            & 86.97 & 86.32          & 88.41  & 95.96          & 89.43          & 95.97 & 96.62      & 91.92          & \textbf{98.37} \\
		AA(\%)            & 74.72 & 71.18          & 75.77  & 89.45          & 79.12          & 91.74 & 93.10      & 91.84          & \textbf{96.15} \\
		$\kappa\times100$ & 84.71 & 83.88          & 86.40  & 95.27          & 87.62          & 95.29 & 96.05      & 90.61          & \textbf{98.09} \\  \bottomrule
	\end{tabular}
}
\end{table*}

\begin{table*}[]
	\centering
	\caption{\modified{Classification results of different methods on the Houston 2018 dataset with 10\% training samples\label{tab11}}}
	\resizebox{\linewidth}{!}{
		\begin{tabular}{c|ccccccccc}
			\toprule
			No.              & SVM\cite{1323134} & 2D-CNN\cite{7450160}    & 3D-CNN\cite{chen2016deep}     & HybridSN\cite{8736016} & SpectralFormer\cite{9627165} & SSFTT\cite{9684381}           & MASSFormer\cite{10506482}      & SS-Mamba\cite{rs16132449}        & Ours            \\ \hline
			1                 & 83.25           & 83.41           & 89.77  & 85.96           & 87.01          & 84.05           & \textbf{91.87}  & 91.84           & 83.01           \\
			2                 & 92.30           & 85.70           & 88.65  & \textbf{95.96}  & 94.87          & 94.50           & 95.25           & 83.94           & 95.72           \\
			3                 & 99.35           & \textbf{100.00} & 99.68  & \textbf{100.00} & 98.86          & \textbf{100.00} & \textbf{100.00} & \textbf{100.00} & \textbf{100.00} \\
			4                 & 95.96           & 97.88           & 97.63  & 98.47           & 97.97          & 97.50           & 97.66           & \textbf{98.66}  & 98.15           \\
			5                 & 69.32           & 53.78           & 75.91  & 94.09           & 87.23          & 96.72           & 92.27           & 86.57           & \textbf{97.04}  \\
			6                 & 97.69           & 92.45           & 85.97  & \textbf{100.00} & 98.62          & \textbf{100.00} & 99.85           & 97.27           & \textbf{100.00} \\
			7                 & \textbf{100.00} & 98.33           & 98.33  & 99.58           & 59.83          & 97.49           & 99.58           & 99.84           & \textbf{100.00} \\
			8                 & 87.38           & 84.69           & 88.88  & 97.02           & 92.39          & 98.83           & 97.10           & 86.86           & \textbf{99.68}  \\
			9                 & 95.53           & 99.33           & 96.25  & 99.23           & 98.03          & 99.34           & \textbf{99.70}  & 64.52           & 99.67           \\
			10                & 51.34           & 57.29           & 57.39  & 90.86           & 76.35          & 92.98           & \textbf{95.70}  & 44.07           & 94.89           \\
			11                & 48.00           & 59.88           & 52.30  & 85.79           & 70.98          & 90.13           & \textbf{91.83}  & 41.42           & 89.61           \\
			12                & 23.02           & 20.89           & 10.16  & 35.42           & \textbf{81.91} & 54.09           & 73.07           & 65.64           & 68.18           \\
			13                & 79.18           & 74.12           & 80.82  & 95.32           & 86.59          & 96.18           & 95.70           & 50.22           & \textbf{98.37}  \\
			14                & 71.86           & 73.25           & 89.20  & 97.98           & 88.13          & 96.53           & \textbf{99.28}  & 95.55           & 99.25           \\
			15                & 89.12           & 73.84           & 91.48  & \textbf{99.70}  & 99.14          & \textbf{99.70}  & 99.43           & 99.00           & \textbf{99.70}  \\
			16                & 74.81           & 73.56           & 88.65  & 98.31           & 94.57          & 98.97           & 98.90           & 94.00           & \textbf{99.49}  \\
			17                & 30.57           & \textbf{100.00} & 96.55  & 76.72           & 96.74          & 96.55           & \textbf{100.00} & \textbf{100.00} & \textbf{100.00} \\
			18                & 50.00           & 69.19           & 83.89  & 94.23           & 78.49          & 95.57           & \textbf{99.69}  & 92.29           & 98.14           \\
			19                & 83.87           & 98.15           & 97.12  & 98.97           & 88.63          & 99.40           & 99.66           & 98.12           & \textbf{100.00} \\
			20                & 85.40           & 95.23           & 96.11  & \textbf{99.98}  & 98.60          & \textbf{99.98}  & \textbf{99.98}  & 98.58           & 99.92           \\ \hline
			OA(\%)            & 84.11           & 86.11           & 86.68  & 96.32           & 91.70          & 97.00           & 97.64           & 68.06           & \textbf{97.72}  \\
			AA(\%)            & 72.72           & 80.54           & 82.73  & 92.18           & 79.91          & 94.43           & 94.61           & 84.42           & \textbf{95.04}  \\
			$\kappa\times100$ & 78.73           & 81.06           & 82.34  & 95.22           & 89.00          & 96.05           & 96.90           & 61.49           & \textbf{96.99}  \\ \bottomrule
		\end{tabular}
	}
\end{table*}
 
\modified{4) WHU-Hi-HanChuan: The classification results for the WHU-Hi-HanChuan dataset are shown in Table~\ref{tab10}. Our method achieves the highest classification accuracy for all three metrics: OA  of 98.37\%, AA  of 96.15\%, and $\kappa \times 100$ of 98.09\%. Compared to the second-place method, MASSFormer, our method improves OA, AA, and $\kappa$ by 1.75\%, 3.05\%, and 2.04\%, respectively. When compared to traditional 2D convolution methods, OA shows a significant improvement of 12.05\%. Moreover, our method achieves the highest classification accuracy in 12 out of the 16 categories in the dataset. The classification maps in Figs.~\ref{fig8} clearly show that our classification results are closer to the ground truth categories, especially in the tree region, highlighting the model's advantage in complex scenes.}

\modified{5) Houston2013: As shown in Table~\ref{tab11}, the Houston2018 dataset includes 20 categories, significantly increasing the complexity of the classification task. SS-Mamba performs poorly in feature fusion, leading to information loss in complex environments and resulting in subpar classification performance. In contrast, our CFF module effectively reduces information loss between different layers of encoders, achieving the highest classification accuracy with OA, AA, and  $\kappa \times 100$ values of 97.72\%, 95.04\%, and 96.99\%, respectively. Compared to SS-Mamba, OA is improved by 29.66\%, demonstrating the superior performance of our model. The classification results shown in Figs.~\ref{fig9} indicate that our method achieves a significantly higher match with the ground truth categories compared to other models, especially in complex scenes.}


\subsection{Ablation Study}

\begin{table}[]
\caption{Ablation studies of the proposed TBFE, HPA, and CFF modules on the Salinas dataset\label{tab12}}
\resizebox{\linewidth}{!}{
\begin{tabular}{c|ccc|ccc}
\toprule
\multirow{2}{*}{Cases} & \multicolumn{3}{c|}{Component} & \multicolumn{3}{c}{Metrics}                   \\ \cline{2-7} 
                       & TBFE      & HPA        & CFF       & OA(\%)         & AA(\%)         & $\kappa\times 100$      \\ \hline
1                      & 3D+2D      & \ding{55} & \ding{55} & 94.69          & 96.33          & 94.09      \\
2                     & \ding{51}  & \ding{55} & \ding{55} & 97.35          & 98.34          & 97.05      \\
3                      & \ding{51}  & \ding{55} & \ding{51} & 98.00          & 98.19          & 97.77      \\
4                     & \ding{51}  & \ding{51} & \ding{55} & 97.62          & 98.55          & 97.35      \\ 
5                      & 3D+2D      & \ding{51} & \ding{51} & 97.61          & 98.48          & 97.35      \\
6                      & \ding{51}  & \ding{51} & \ding{51} & \textbf{98.67} & \textbf{98.64} & \textbf{98.52} \\ \bottomrule
\end{tabular}
}
\end{table}

\begin{table*}[]
	\centering
	\caption{Comparison of testing times, trainable parameters, and OA of different approaches on Pavia University dataset\label{tab13}}
	\resizebox{\linewidth}{!}{
		\begin{tabular}{c|cccccccc}
			\toprule
			& 3D-CNN\cite{chen2016deep}        & HybridSN\cite{8736016} & GAHT\cite{9895238}   & SpectralFormer\cite{9627165} & SSFTT\cite{9684381}  & \modified{MASSFormer\cite{10506482}} & \modified{SS-Mamba\cite{rs16132449}} & Ours           \\ \hline
			Testing Time(s) & \textbf{6.39} & 7.29     & 13.69  & 18.52          & 7.24   & 10.73      & 10.41    & 8.02           \\
			Params(k)       & 462.486       & 797.57   & 946.83 & \textbf{128.8} & 148.49 & 314.41     & 470.0    & 167.49         \\
			OA(\%)          & 92.01         & 98.16    & 96.40  & 93.17          & 99.70  & 99.48      & 95.79    & \textbf{99.92} \\ \bottomrule
		\end{tabular}
	}
\end{table*}

To demonstrate the effectiveness of the proposed method, we conducted ablation experiments on the Salinas dataset to assess the impact of each component on classification performance. The primary components of the proposed method include TBFE, HPA, and CFF. The experimental results are presented in Table~\ref{tab12}.

Specifically, in Case 1, we replaced the TBFE block with concatenated Conv3D and Conv2D layers to evaluate the performance of the baseline without the HPA and CFF modules. Under this configuration, the cross-layer connections between the encoders were removed, and the encoders were \modified{connected directly} in series. In Case 2, we added the TBFE module to the baseline established in Case 1 to assess the impact of dual-branch convolution on the model's classification performance. \modified{The inclusion of} the TBFE block resulted in a 2.66\% improvement in OA for Case 2. Case 3 introduced the CFF block to explore the effect of cross-layer feature fusion on classification results. With the CFF block incorporated, Case 3 achieved an OA of 98.00\%. In Case 4, the HPA block was introduced to the baseline established in Case 2 to investigate the effect of multi-layer channel attention on classification results. The comparative analysis demonstrated that Case 4 achieved higher OA, AA, and $\kappa$ scores compared to Case 2, indicating that multi-layer channel attention enhances the spatial feature representation capabilities of the data. Finally, by comparing Case 5 and Case 6, we investigated the impact of TBFE on the model's classification performance in the presence of HPA and CFF. The experimental results highlighted the strong feature-capturing capabilities of the dual-branch convolution module. In summary, each component of the proposed method plays a beneficial role in enhancing classification performance.

\subsection{Model Complexity and Efficiency Analysis}

We evaluated the computational performance of several classical HSI classification methods on the Pavia University dataset, utilizing a training sample rate of 5\%. As shown in Table~\ref{tab13}, our proposed method demonstrates notable improvement in both testing time and parameter size, while simultaneously attaining state-of-the-art classification performance.

In terms of testing time, our method demonstrates relatively faster execution compared to several state-of-the-art methods. The 3D-CNN and HybridSN methods, which rely solely on convolutional neural networks, exhibit quicker runtimes. SSFTT accelerates runtime by significantly reducing spectral dimensions through a tokenizer. \modified{Although the Mamba model delivers faster inference speed, the SS-Mamba model requires more time due to the incorporation of spatial feature fusion. Comparatively, our method shows a modest advantage in testing time. However, the additional integration of convolutional extraction features and the enhancement of information interaction through skip connections results in slightly more time compared to pure convolutional models and simpler Transformer architectures.}

In terms of model parameter size, our proposed method also outperforms most of the comparison approaches. The structure of the SpectralFormer network is relatively simple, while SSFTT merely combines convolution and transformer techniques, both relying on a single scale during the feature extraction process. \modified{MASSFormer results in additional computational overhead, due to the need for memory tokens to store prior knowledge.} In contrast, our method stacks multiple Transformer encoders to fully leverage \modified{both}  spectral and spatial information in HSIs, leading to improved classification accuracy with only a slight increase in parameter size.

In summary, our proposed method achieves leading accuracy with a moderate parameter size, while effectively minimizing computational overhead. This further demonstrates the advantages of our approach.

\section{Conclusion}
In this work, we propose a synergistic CNN-Transformer network with pooling attention fusion for HSI classification, presenting a novel and efficient approach to this task. Our method effectively integrates shallow CNNs with transformers, enabling robust spatial feature extraction while accounting for long-range dependencies between spectral sequence properties and spatial features. We also developed three efficient modules: TBFE, HPA, and CFF. The TBFE module captures shallow features through dual-branch convolution, which are subsequently fused by the HPA module across various dimensions to create a more comprehensive feature representation. These extracted features are then fed into the transformer encoder, where cross-layer feature fusion mitigates information loss between layers. Extensive experiments confirm that our method exceeds the state-of-the-art methods and obtains satisfactory results.

In the future, leveraging multimodal knowledge may improve the proposed architecture beyond solely vision modality. Future research will explore enhancing the model's ability to encode multimodal features to achieve a richer representation of information. \modified{Furthermore, as the model depth increases, both computational complexity and runtime will rise significantly. Therefore, we will continue to explore more effective feature dimensionality reduction methods to improve both computational efficiency and inference speed.}

% \section*{Funding}
% This work was supported in part by the Natural Science Foundation of Guangdong Province under Grant 2024A1515011766, State key laboratory major special projects of Jilin Province Science and Technology Development Plan under Grant SKL202402024, and the Basic and Frontier Research Programmes of Chongqing under Grant CSTB2022NSCQ-MSX0583.

\bibliographystyle{elsarticle-num} 
\bibliography{references}

\end{document}